\documentclass[11pt]{article}

\usepackage{microtype} %
\usepackage{booktabs}  %
\usepackage{url}  %
\usepackage{csquotes}
\usepackage{graphicx} 
\usepackage{amsmath}
\usepackage{amsthm}
\usepackage{tabularx} 
\usepackage{multirow}
\usepackage[inline]{enumitem}

\usepackage[shortpaper,final]{automl}
\usepackage{cleveref}

\usepackage{natbib}
\title{Quickly Tuning Foundation Models for Image Segmentation}

\author[1]{\nameemail{Breenda Das}{dasb@informatik.uni-freiburg.de}}
\author[1]{\nameemail{Lennart Purucker}{purucker@cs.uni-freiburg.de}}
\author[2,1]{\nameemail{Timur Carstensen}{carstent@cs.uni-freiburg.de}}
\author[3,2,1]{\nameemail{Frank Hutter}{fh@cs.uni-freiburg.de}}

\affil[1]{University of Freiburg}
\affil[2]{ELLIS Institute Tübingen}
\affil[3]{Prior Labs}

\hypersetup{%
  pdfauthor={AutoML}, %
  pdftitle={Documentation for the automl Package},
  pdfsubject={Documentation for automl Package},
  pdfkeywords={AutoML, LaTeX, style}
}

\begin{document}

\maketitle

\begin{abstract}
Foundation models like SAM (Segment Anything Model) exhibit strong zero-shot image segmentation performance, but often fall short on domain-specific tasks. Fine-tuning these models typically requires significant manual effort and domain expertise. In this work, we introduce QTT-SEG, a meta-learning-driven approach for automating and accelerating the fine-tuning of SAM for image segmentation. Built on the Quick-Tune hyperparameter optimization framework, QTT-SEG predicts high-performing configurations using meta-learned cost and performance models, efficiently navigating a search space of over 200 million possibilities. We evaluate QTT-SEG on eight binary and five multiclass segmentation datasets under tight time constraints. Our results show that QTT-SEG consistently improves upon SAM’s zero-shot performance and surpasses AutoGluon Multimodal, a strong AutoML baseline, on most binary tasks within three minutes. On multiclass datasets, QTT-SEG delivers consistent gains as well. These findings highlight the promise of meta-learning in automating model adaptation for specialized segmentation tasks. Code available at: \href{https://github.com/ds-brx/QTT-SEG}{Link.}
\end{abstract}

\begin{figure}
  \centering
  \includegraphics[width=1\linewidth]{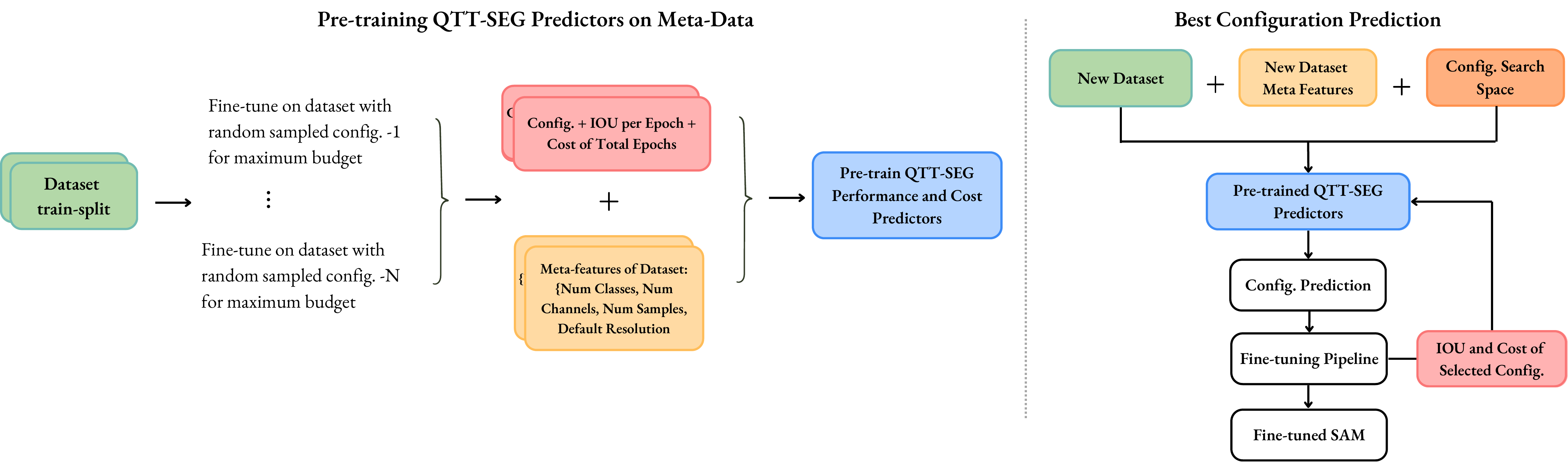}
  \caption{\textbf{Overview of QTT-SEG}: It first pre-trains performance and cost predictors using dataset meta-features, performance and cost traces from multiple configurations. These predictors are then used to guide efficient pipeline selection and fine-tuning on new datasets.}
  \label{fig:flow}
\end{figure}

\begin{figure}[ht]
  \centering
  \begin{subfigure}[b]{0.45\linewidth}
    \centering
    \includegraphics[width=\linewidth]{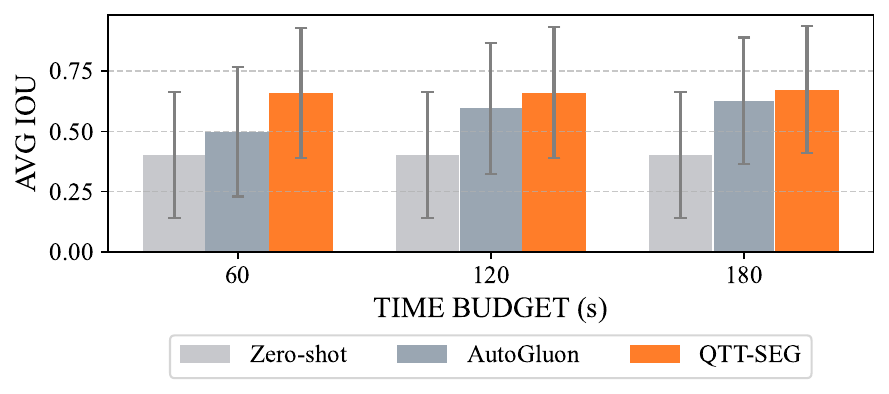}
    \caption{Binary Segmentation}
    \label{fig:Binary_datasets}
  \end{subfigure}
  \hfill
  \begin{subfigure}[b]{0.45\linewidth}
    \centering
    \includegraphics[width=\linewidth]{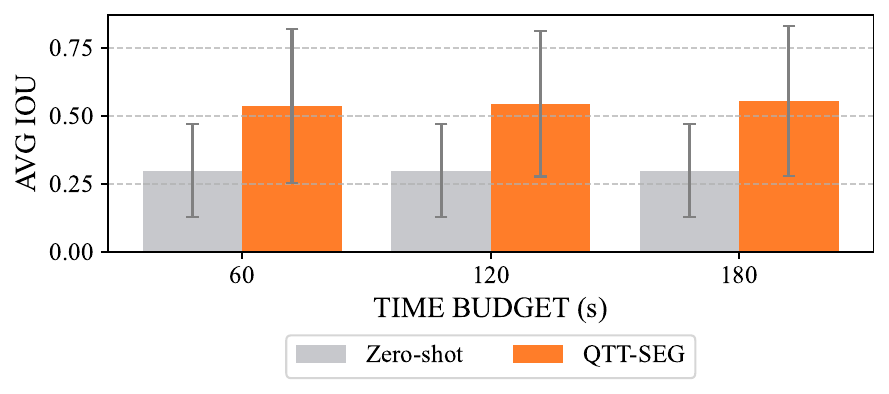}
    \caption{Multiclass Segmentation}
    \label{fig:multi_datasets}
  \end{subfigure}
    \caption{
        \textbf{Performance over Time Budgets:} Mean IoU (bars) and standard deviation (error bars) of Zero-shot, AG(Autogluon-Multimodal), and QTT-SEG across all binary and multiclass segmentation tasks. QTT-SEG shows consistent gains with longer budgets. \textit{Note: AG is evaluated only on binary tasks due to missing out-of-the-box multiclass support.}
    }

  \label{fig:avg_performance}
\end{figure}

\begin{table*}[ht]
\centering
\caption{Binary Semantic Segmentation ranked by QTT-SEG gain over Zero-shot in 60s. Values represent mean accuracy with standard deviation subscripted.}
\label{tab:segmentation_results_sorted}
\begin{scriptsize}
\makebox[\textwidth]{%
\begin{tabular}{l|c|cc|cc|cc}
\toprule
\textbf{Dataset} 
& \textbf{Zero-shot} 
& \multicolumn{2}{c|}{\textbf{60 SEC}} 
& \multicolumn{2}{c|}{\textbf{120 SEC}} 
& \multicolumn{2}{c}{\textbf{180 SEC}} \\
\cmidrule(r){3-4} \cmidrule(r){5-6} \cmidrule(r){7-8}
& & AG & QTT-SEG & AG & QTT-SEG & AG & QTT-SEG \\
\midrule

polyp   & $0.495_{0.006}$ & $0.328_{\text{0.183}}$ & $\mathbf{0.850_{0.031}}$ 
        & $0.622_{0.027}$ & $\mathbf{0.866_{0.008}}$ 
        & $0.711_{0.011}$ & $\mathbf{0.867_{0.006}}$ \\

lesion  & $0.573_{0.028}$ & $0.704_{0.046}$ & $\mathbf{0.870_{0.008}}$ 
        & $0.802_{0.016}$ & $\mathbf{0.863_{0.009}}$ 
        & $0.806_{0.013}$ & $\mathbf{0.882_{0.008}}$ \\

leaf    & $0.377_{0.010}$ & $0.501_{0.024}$ & $\mathbf{0.657_{0.009}}$ 
        & $0.587_{0.032}$ & $\mathbf{0.635_{0.034}}$ 
        & $0.633_{0.382}$ & $\mathbf{0.646_{0.028}}$ \\

covid   & $0.342_{0.005}$ & $0.397_{0.026}$ & $\mathbf{0.612_{0.018}}$ 
        & $0.447_{0.029}$ & $\mathbf{0.647_{0.013}}$ 
        & $0.499_{0.021}$ & $\mathbf{0.636_{0.017}}$ \\

eyes    & $0.069_{0.002}$ & $0.216_{\text{0.102}}$ & $\mathbf{0.291_{0.028}}$ 
        & $\mathbf{0.323_{\text{0.025}}}$ & $0.314_{0.023}$ 
        & $\mathbf{0.365_{0.013}}$ & $0.323_{0.012}$ \\

fiber   & $0.007_{0.000}$ & $0.203_{0.017}$ & $\mathbf{0.229_{0.016}}$ 
        & $\mathbf{0.241_{0.007}}$ & $0.196_{0.111}$ 
        & $0.249_{0.009}$ & $\mathbf{0.250_{0.017}}$ \\

cardiac & $0.764_{0.009}$ & $0.733_{0.040}$ & $\mathbf{0.875_{0.012}}$ 
        & $\mathbf{0.828_{0.009}}$ & $0.866_{0.019}$ 
        & $0.835_{0.007}$ & $\mathbf{0.886_{0.007}}$ \\

chest   & $0.591_{0.002}$ & $\mathbf{0.909_{0.002}}$ & $0.896_{0.010}$ 
        & $\mathbf{0.914_{0.004}}$ & $0.902_{0.004}$ 
        & $\mathbf{0.919_{0.003}}$ & $0.903_{0.014}$ \\

\midrule
\textbf{Average}
        & $0.402$ & $0.499$ & $\mathbf{0.660}$ 
        & $0.595$ & $\mathbf{0.661}$ 
        & $0.627$ & $\mathbf{0.674}$ \\
\bottomrule
\end{tabular}
}
\end{scriptsize}
\vskip -0.1in
\end{table*}

\begin{table*}[ht]
\centering
\caption{Multiclass Semantic Segmentation ranked by QTT-SEG gain over Zero-shot in 60s. Values represent mean accuracy with standard deviation subscripted.}
\label{tab:segmentation_results_multi_sorted}
\begin{scriptsize}
\makebox[\textwidth]{%
\begin{tabular}{l|c|c|c|c}
\toprule
\textbf{Dataset} 
& \textbf{Zero-shot} 
& \textbf{60 SEC (QTT-SEG)} 
& \textbf{120 SEC (QTT-SEG)} 
& \textbf{180 SEC (QTT-SEG)} \\
\midrule

US             & $0.304_{0.002}$ & $0.814_{0.012}$ & $0.784_{0.088}$ & $0.831_{0.013}$ \\
human\_parsing & $0.524_{0.006}$ & $0.843_{0.006}$ & $0.853_{0.008}$ & $0.846_{0.014}$ \\
golf           & $0.151_{0.007}$ & $0.362_{0.047}$ & $0.383_{0.034}$ & $0.391_{0.026}$ \\
terrain        & $0.399_{0.006}$ & $0.466_{0.036}$ & $0.494_{0.030}$ & $0.483_{0.013}$ \\
cholec         & $0.117_{0.007}$ & $0.196_{0.038}$ & $0.216_{0.011}$ & $0.221_{0.018}$ \\

\midrule
\textbf{Average} 
& $0.299$ & $0.536$ & $0.546$ & $0.554$ \\
\bottomrule
\end{tabular}
}
\end{scriptsize}
\vskip -0.1in
\end{table*}

\begin{figure}[ht]
\setlength{\belowcaptionskip}{-5pt}
  \centering
  \resizebox{0.5\linewidth}{!}{ %
    \begin{minipage}{\textwidth} %

    \begin{subfigure}[b]{0.20\linewidth}
      \centering
      \makebox[0pt]{\textbf{Input}}
    \end{subfigure}
    \hfill
    \begin{subfigure}[b]{0.20\linewidth}
      \centering
      \makebox[0pt]{\textbf{Ground Truth}}
    \end{subfigure}
    \hfill
    \begin{subfigure}[b]{0.20\linewidth}
      \centering
      \makebox[0pt]{\textbf{SAM-Zero-Shot}}
    \end{subfigure}
    \hfill
    \begin{subfigure}[b]{0.20\linewidth}
      \centering
      \makebox[0pt]{\textbf{QTT-SEG}}
    \end{subfigure}
    
    \vspace{1mm}
      \centering
      \begin{subfigure}[b]{0.20\linewidth}
        \centering
        \includegraphics[width=\linewidth]{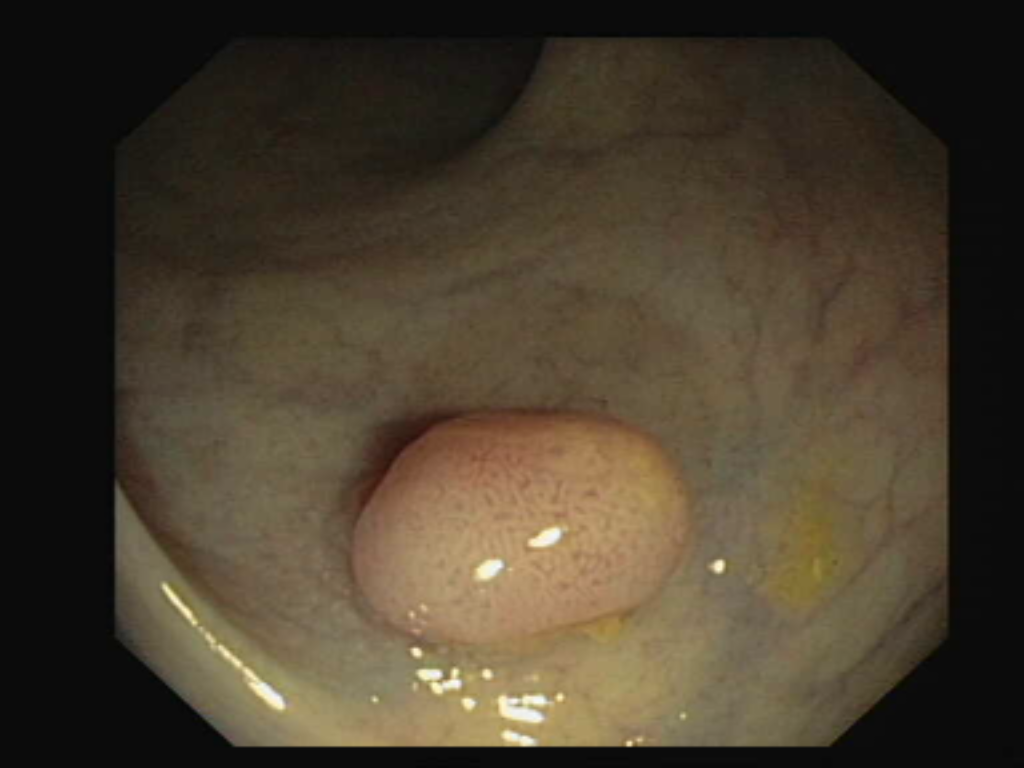}
        \caption*{}
      \end{subfigure}
      \hfill
      \begin{subfigure}[b]{0.20\linewidth}
        \centering
        \includegraphics[width=\linewidth]{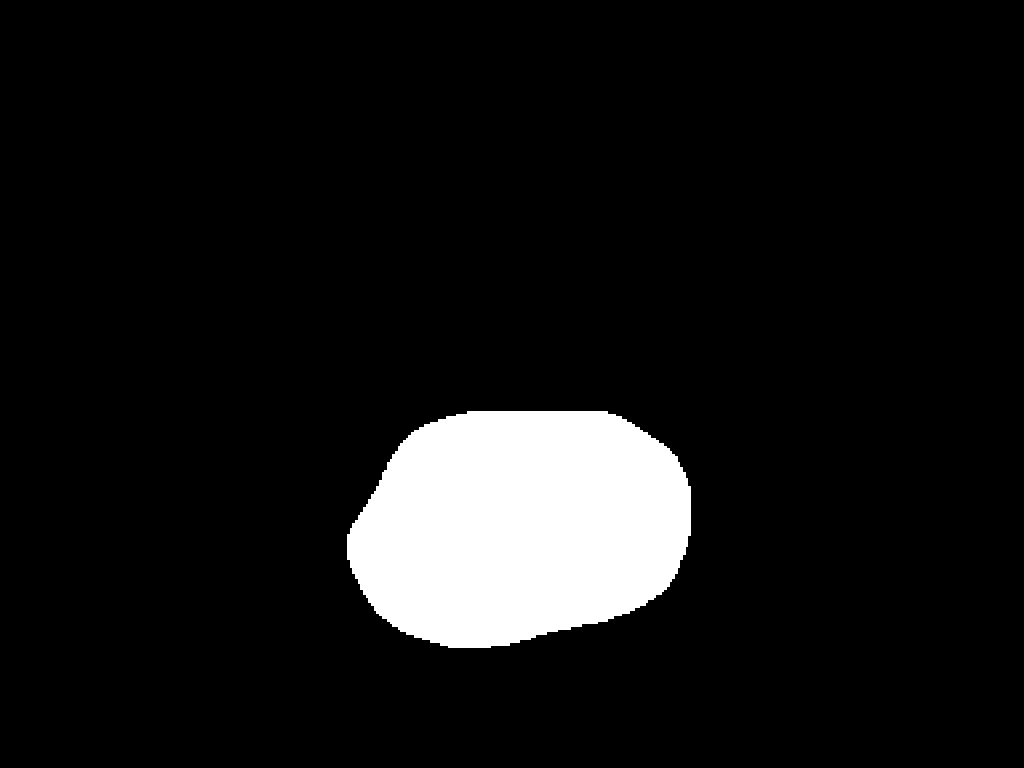}
        \caption*{}
      \end{subfigure}
      \hfill
      \begin{subfigure}[b]{0.20\linewidth}
        \centering
        \includegraphics[width=\linewidth]{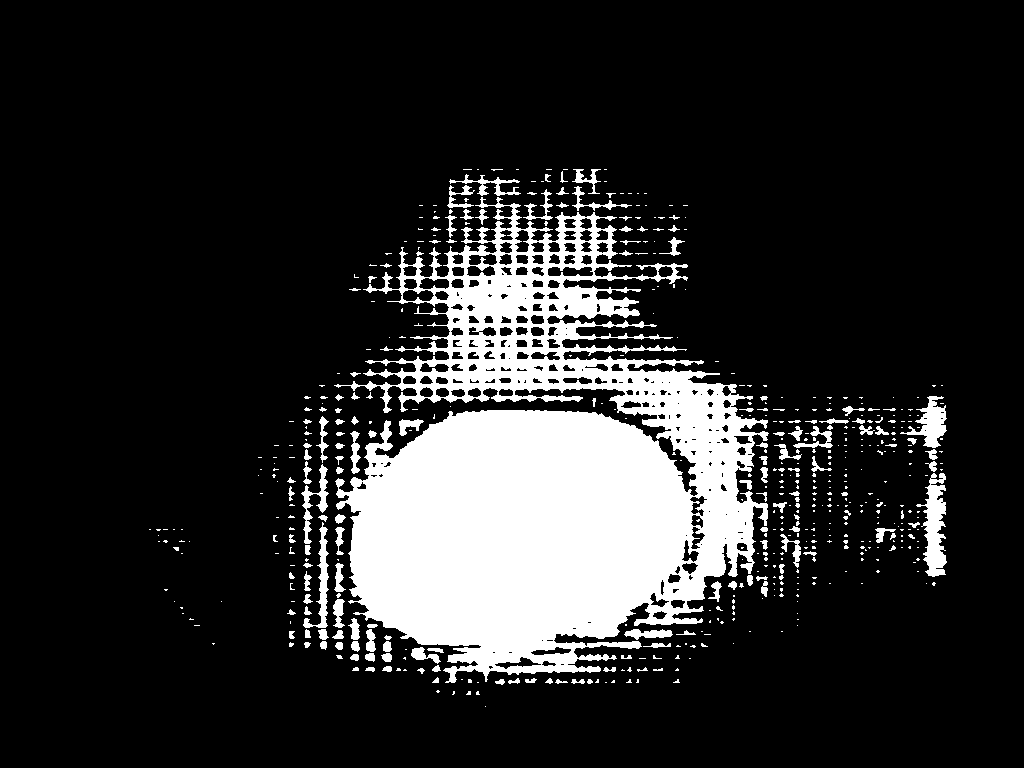}
        \caption*{}
      \end{subfigure}
      \hfill
      \begin{subfigure}[b]{0.20\linewidth}
        \centering
        \includegraphics[width=\linewidth]{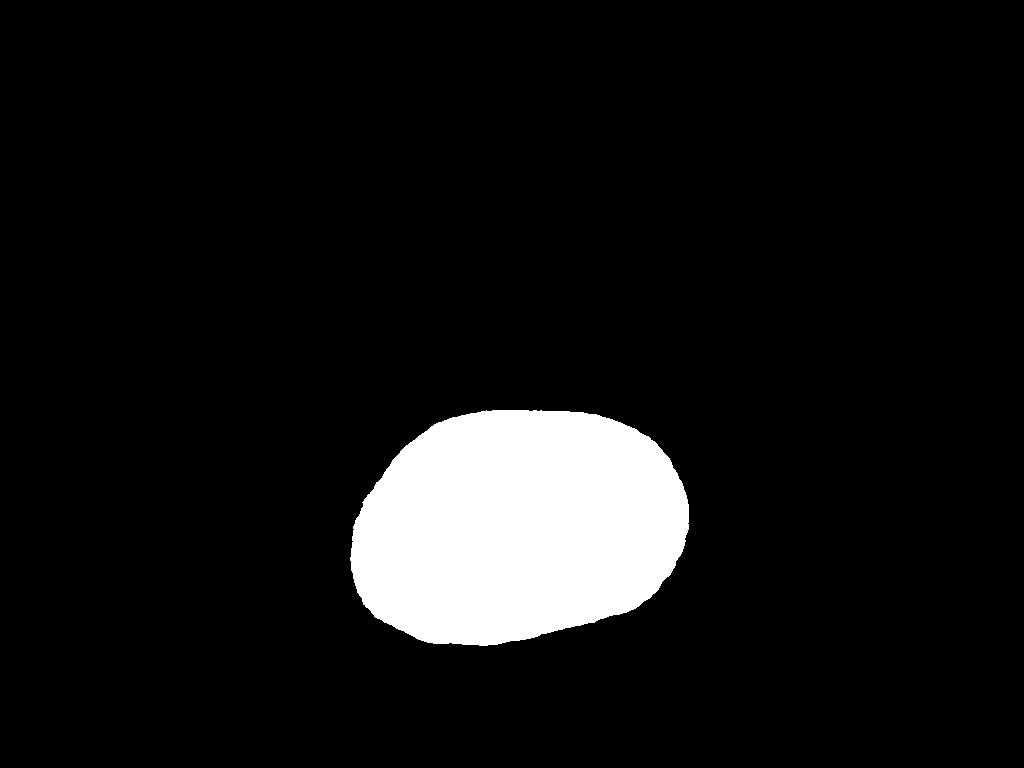}
        \caption*{}
      \end{subfigure}

      \vspace{-4mm}
      \begin{subfigure}[b]{0.20\linewidth}
        \centering
        \includegraphics[width=\linewidth]{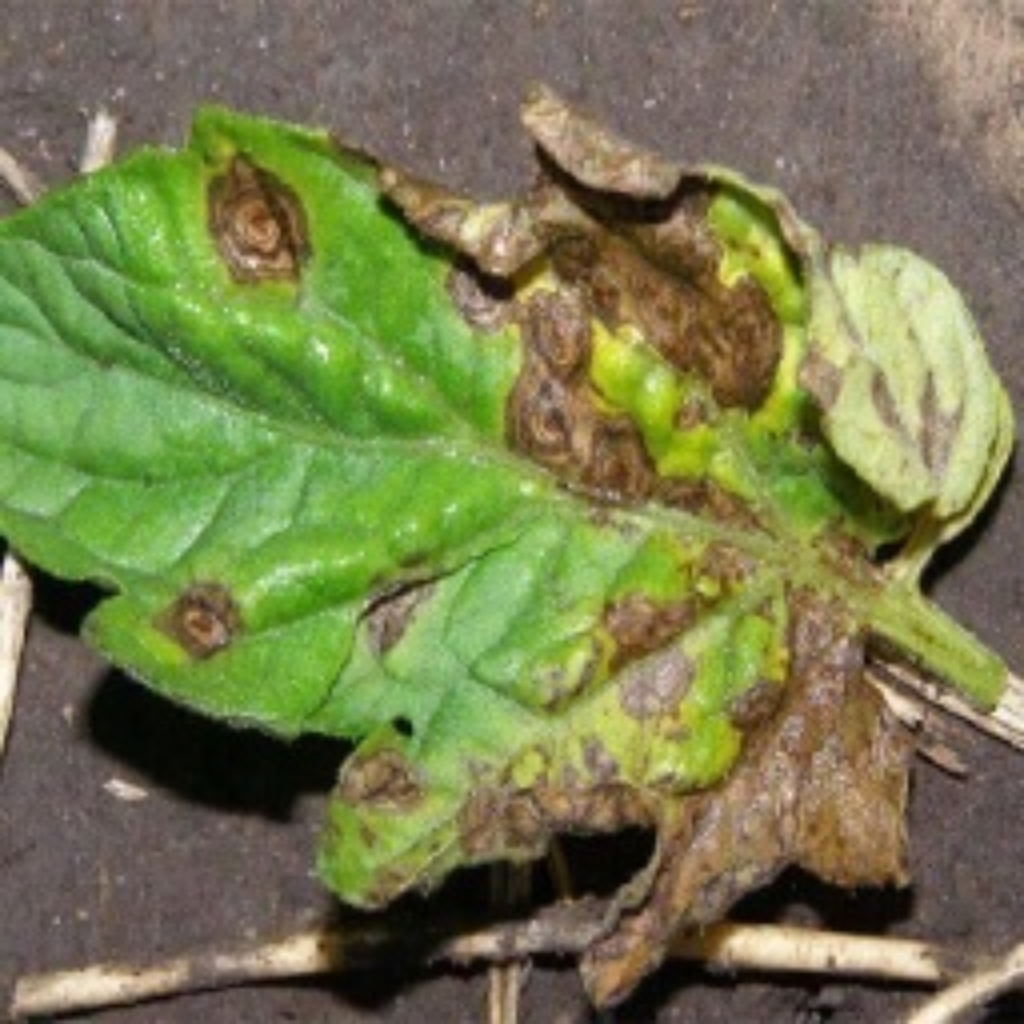}
        \caption*{}
      \end{subfigure}
      \hfill
      \begin{subfigure}[b]{0.20\linewidth}
        \centering
        \includegraphics[width=\linewidth]{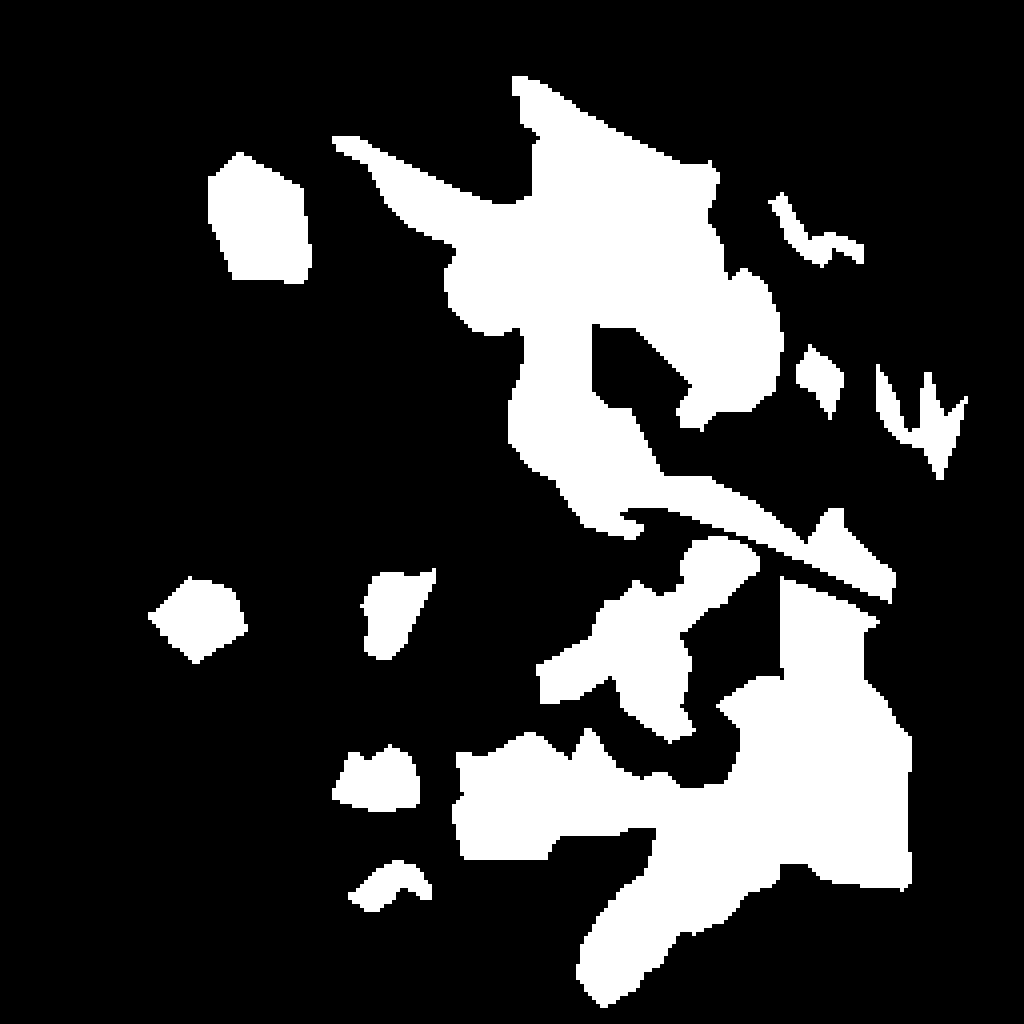}
        \caption*{}
      \end{subfigure}
      \hfill
      \begin{subfigure}[b]{0.20\linewidth}
        \centering
        \includegraphics[width=\linewidth]{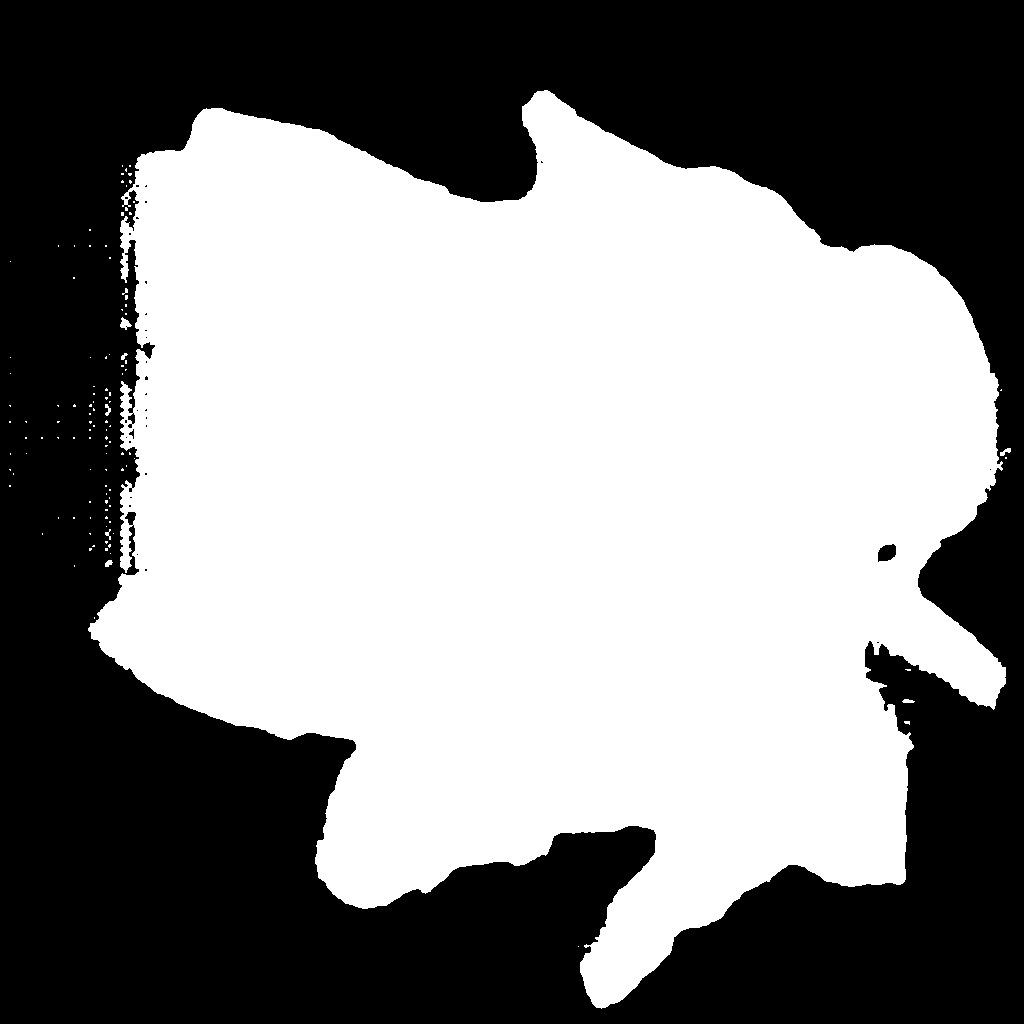}
        \caption*{}
      \end{subfigure}
      \hfill
      \begin{subfigure}[b]{0.20\linewidth}
        \centering
        \includegraphics[width=\linewidth]{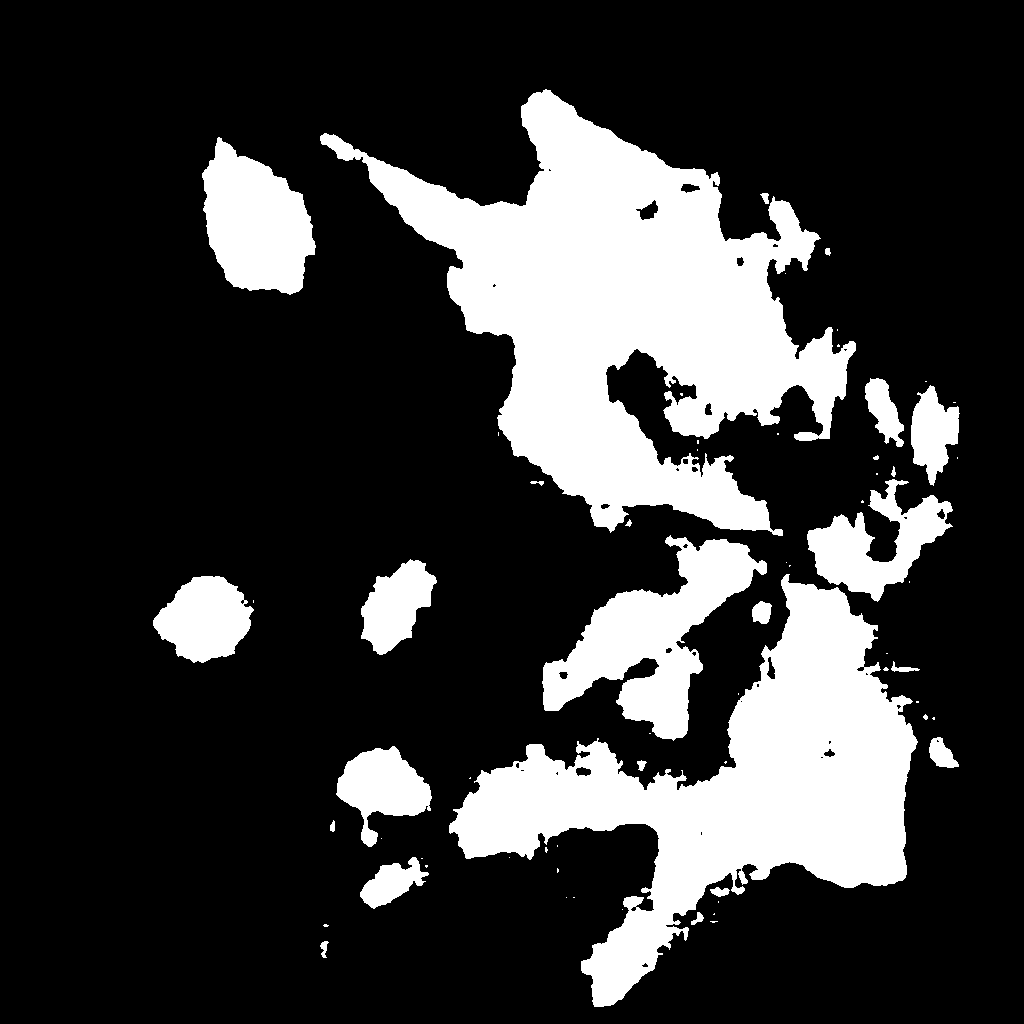}
        \caption*{}
      \end{subfigure}

      \vspace{-4mm}
            \begin{subfigure}[b]{0.20\linewidth}
        \centering
        \includegraphics[width=\linewidth]{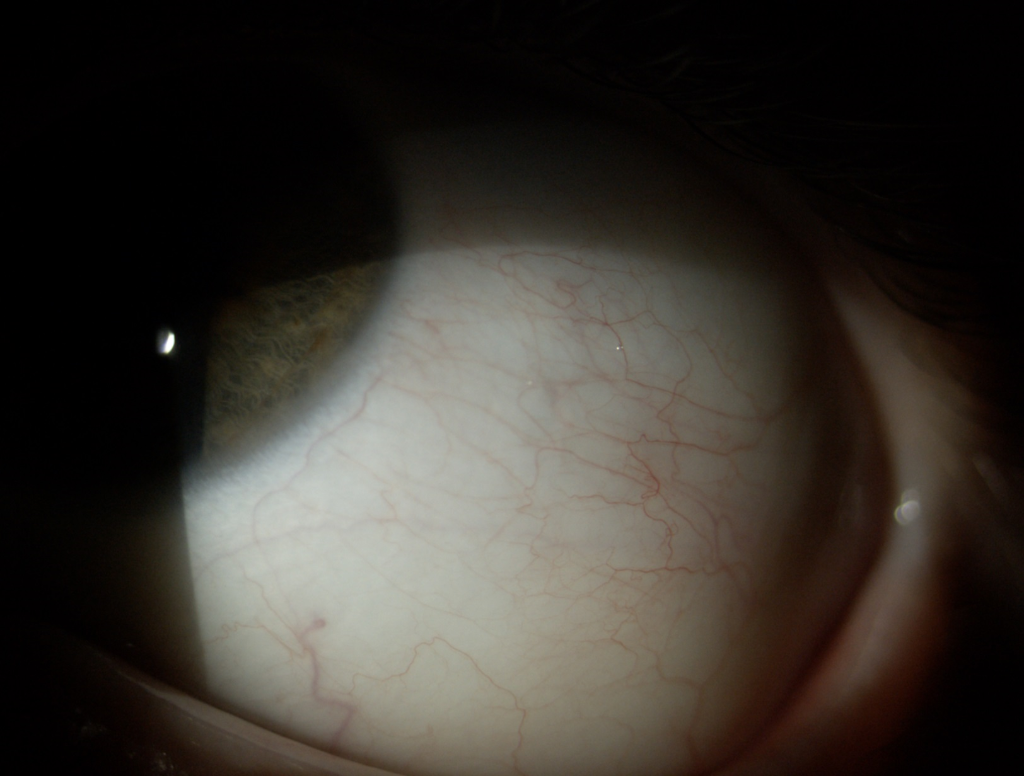}
        \caption*{}
      \end{subfigure}
      \hfill
      \begin{subfigure}[b]{0.20\linewidth}
        \centering
        \includegraphics[width=\linewidth]{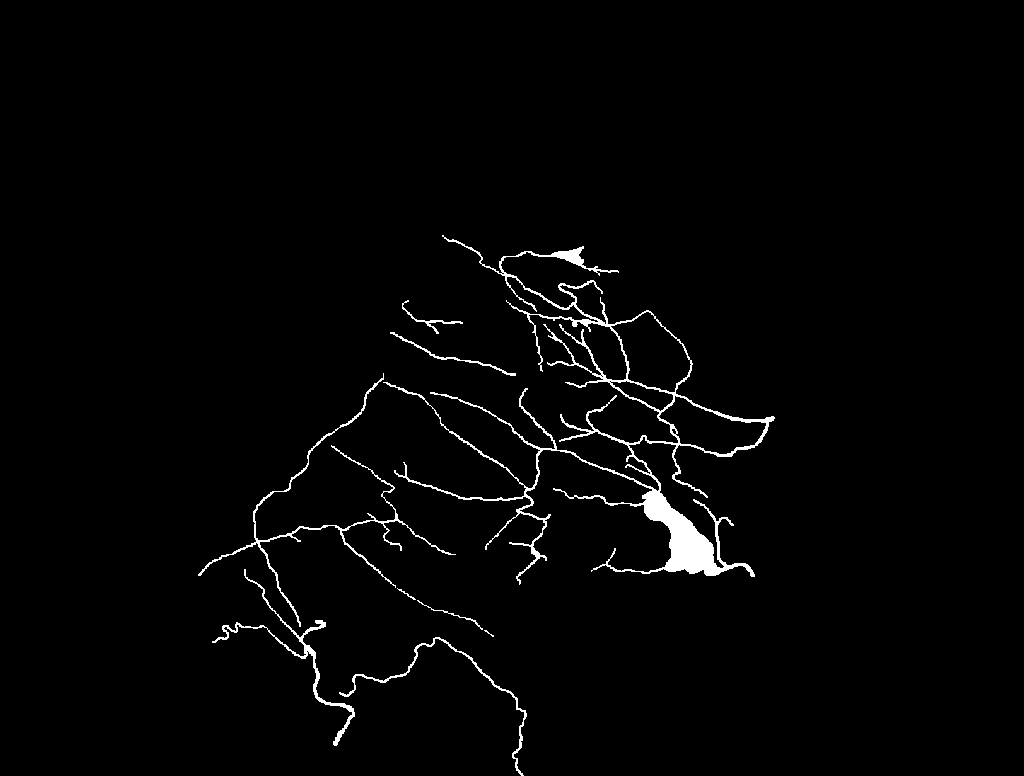}
        \caption*{}
      \end{subfigure}
      \hfill
      \begin{subfigure}[b]{0.20\linewidth}
        \centering
        \includegraphics[width=\linewidth]{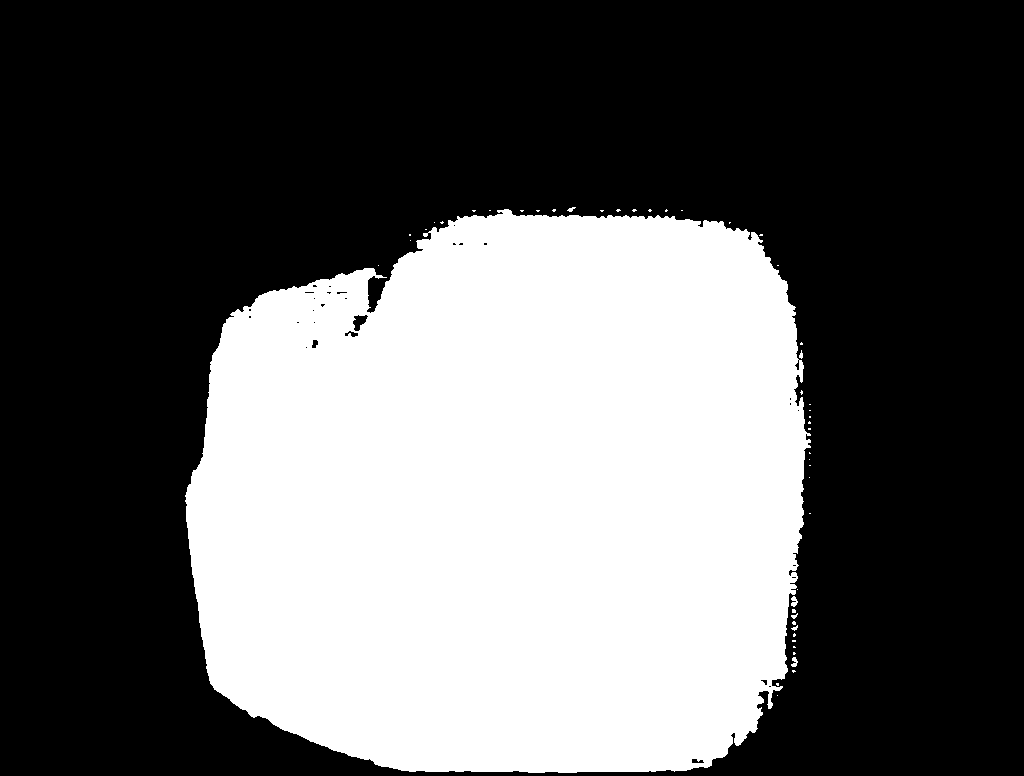}
        \caption*{}
      \end{subfigure}
      \hfill
      \begin{subfigure}[b]{0.20\linewidth}
        \centering
        \includegraphics[width=\linewidth]{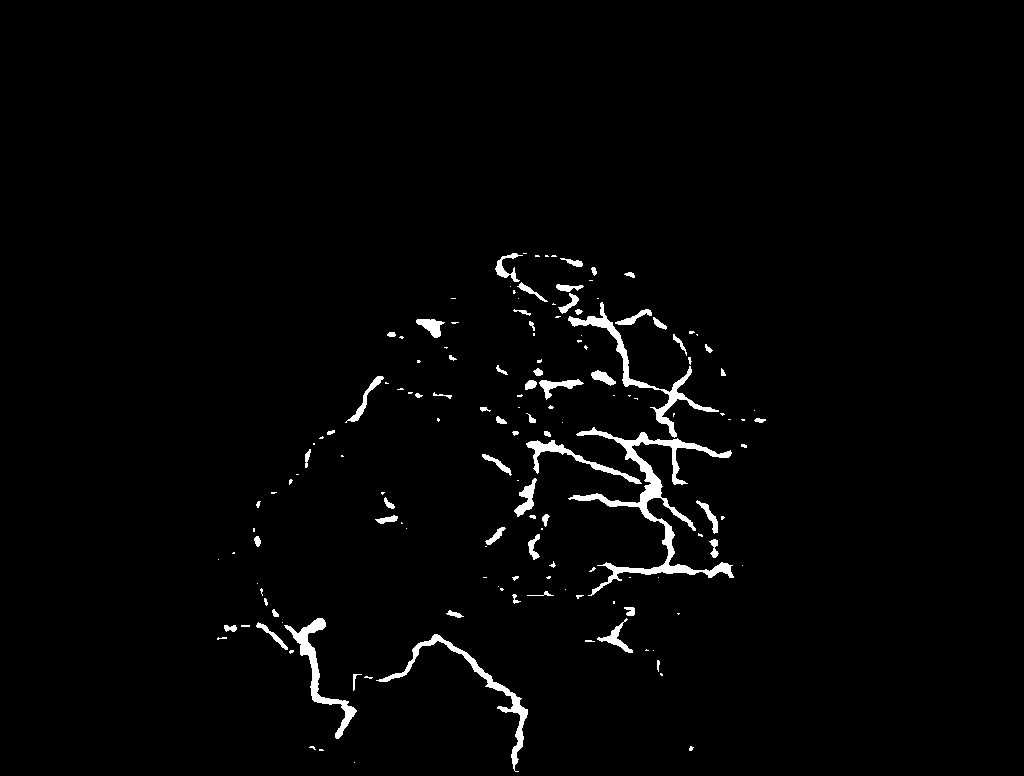}
        \caption*{}
      \end{subfigure}
      
      \vspace{-4mm}
      \begin{subfigure}[b]{0.20\linewidth}
        \centering
        \includegraphics[width=\linewidth]{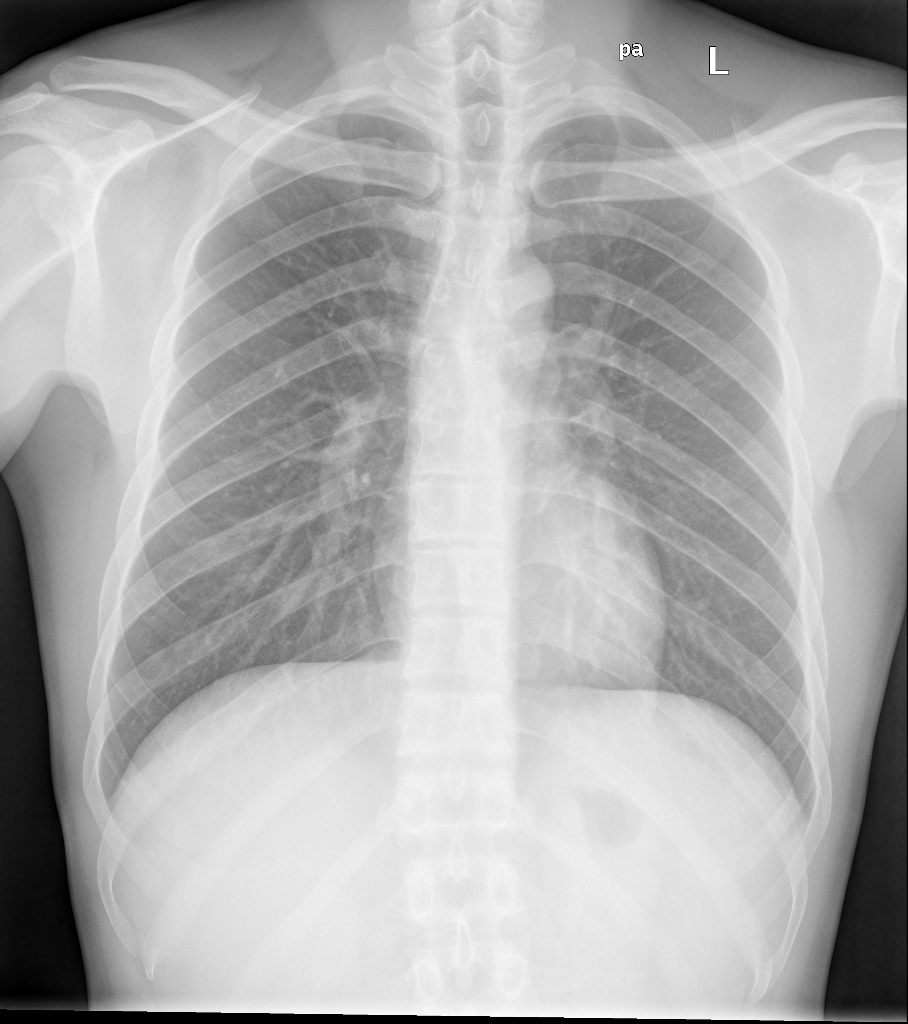}
        \caption*{}
      \end{subfigure}
      \hfill
      \begin{subfigure}[b]{0.20\linewidth}
        \centering
        \includegraphics[width=\linewidth]{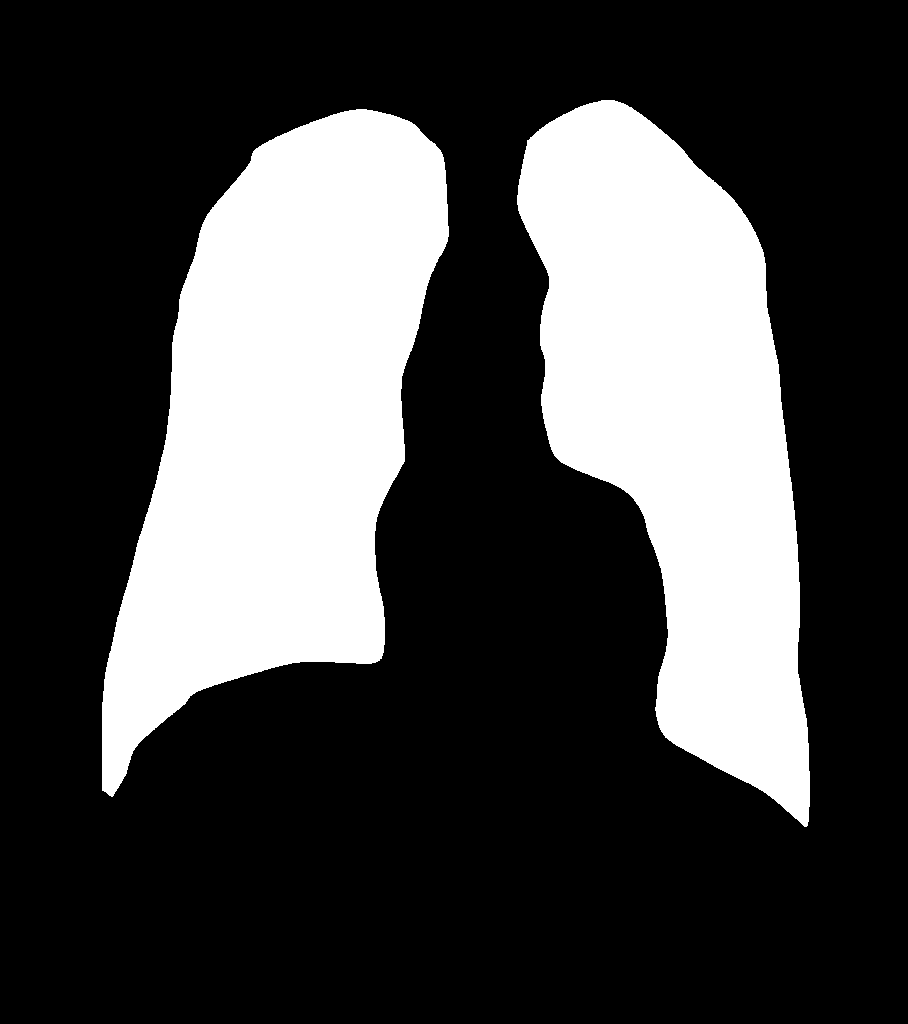}
        \caption*{}
      \end{subfigure}
      \hfill
      \begin{subfigure}[b]{0.20\linewidth}
        \centering
        \includegraphics[width=\linewidth]{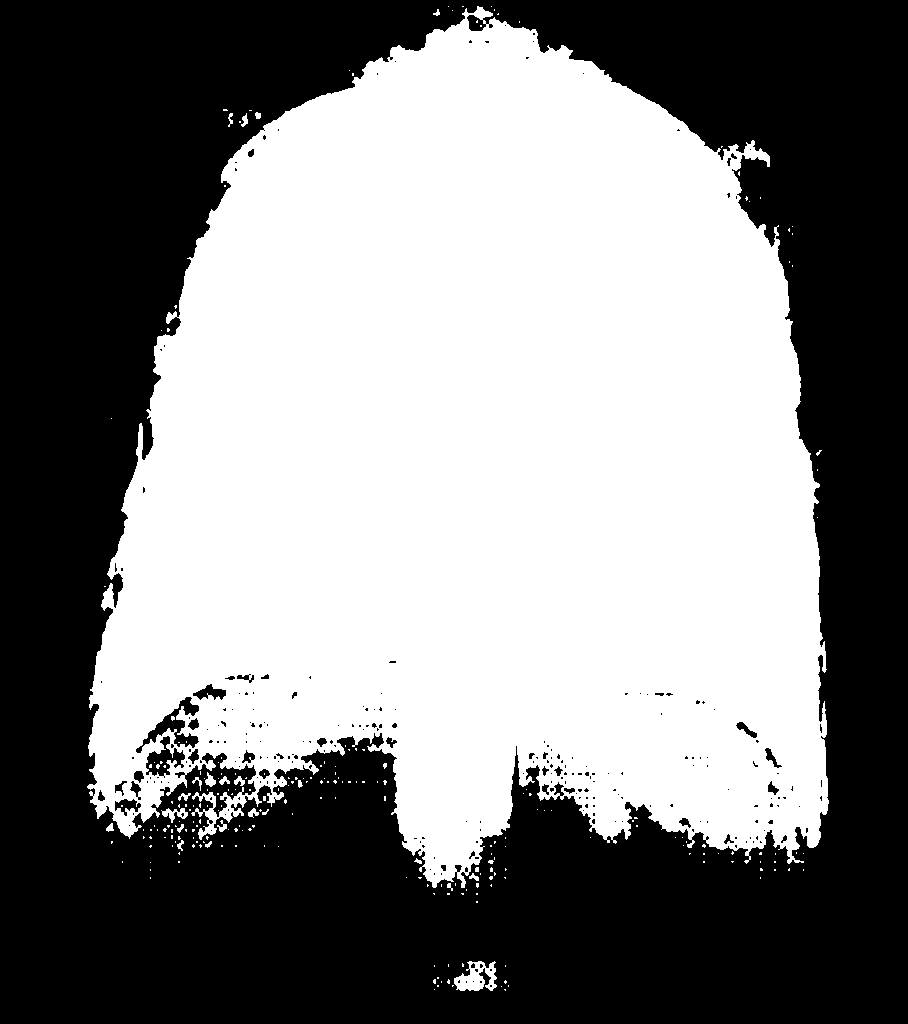}
        \caption*{}
      \end{subfigure}
      \hfill
      \begin{subfigure}[b]{0.20\linewidth}
        \centering
        \includegraphics[width=\linewidth]{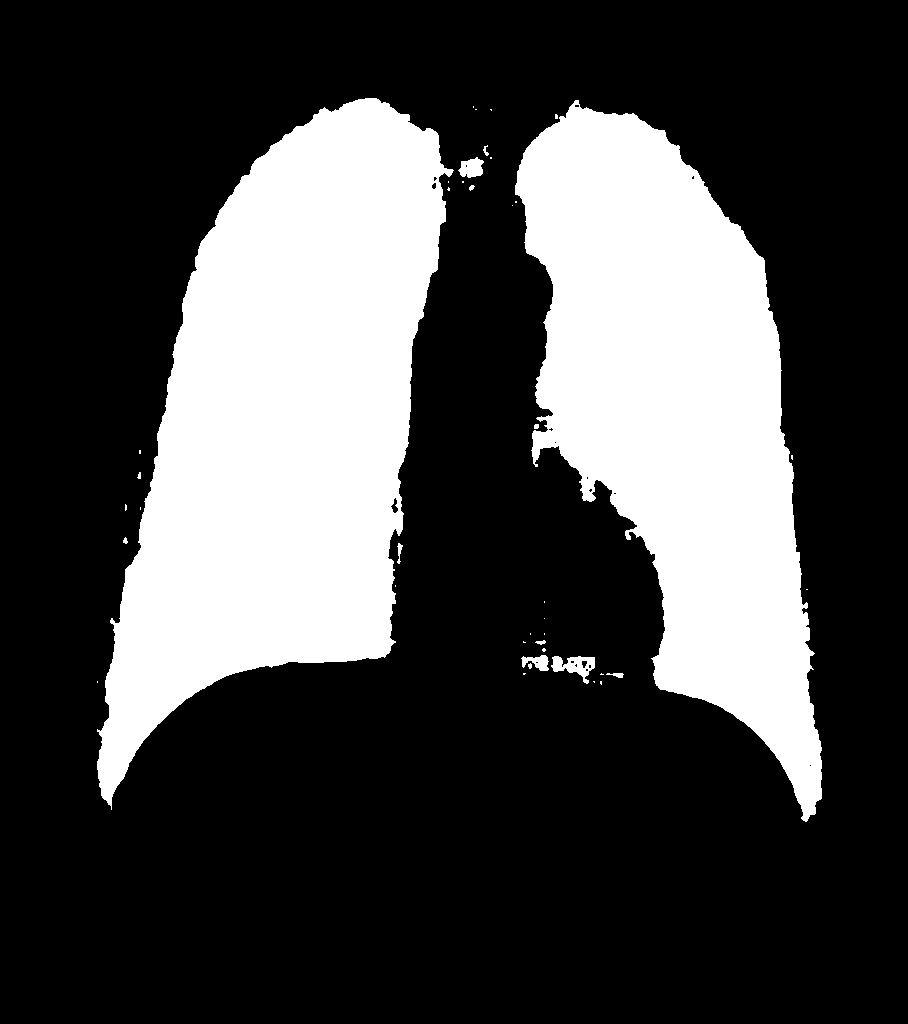}
        \caption*{}
      \end{subfigure}

      \vspace{-4mm}
      \begin{subfigure}[b]{0.20\linewidth}
        \centering
        \includegraphics[width=\linewidth]{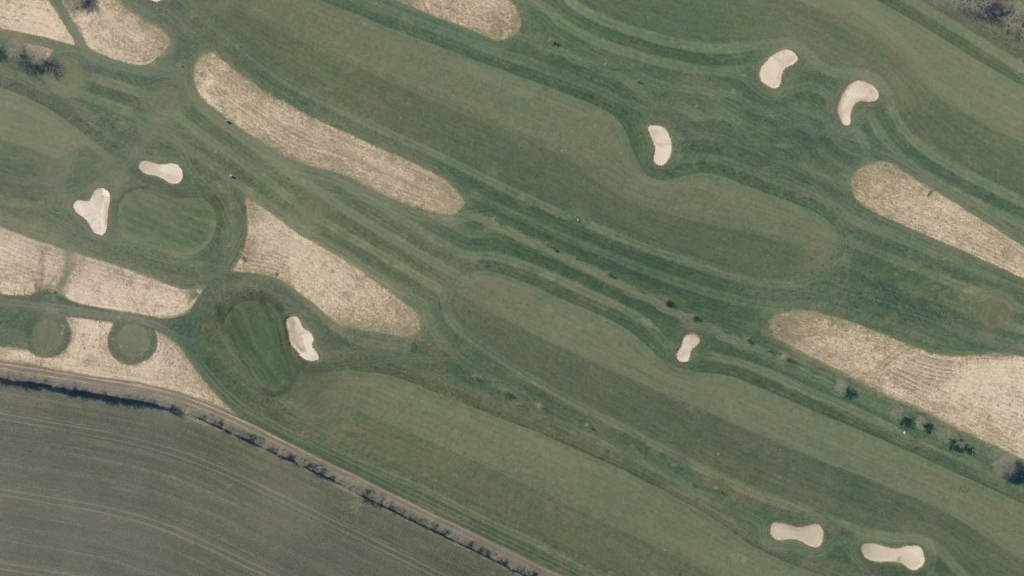}
        \caption*{}
      \end{subfigure}
      \hfill
      \begin{subfigure}[b]{0.20\linewidth}
        \centering
        \includegraphics[width=\linewidth]{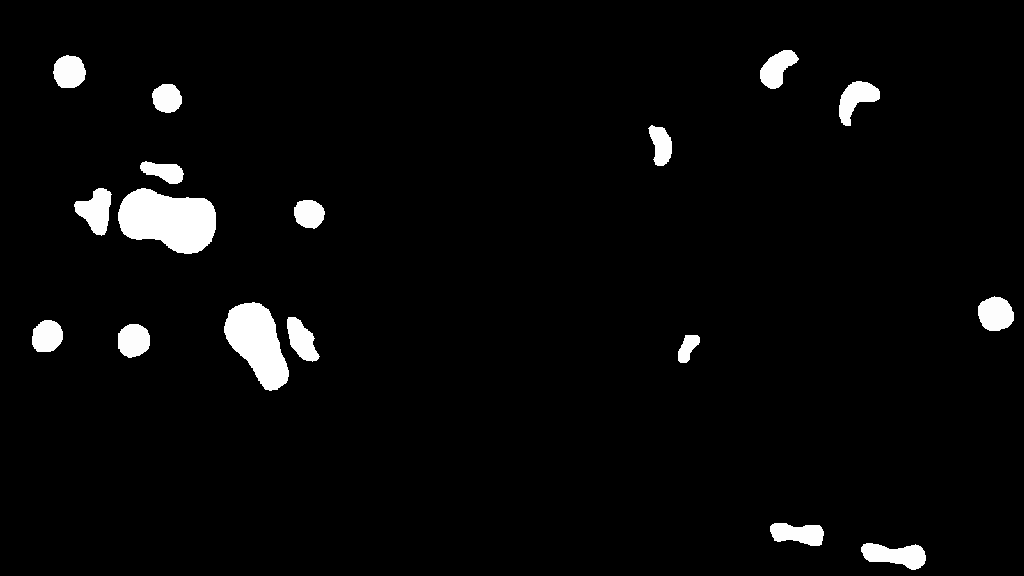}
        \caption*{}
      \end{subfigure}
      \hfill
      \begin{subfigure}[b]{0.20\linewidth}
        \centering
        \includegraphics[width=\linewidth]{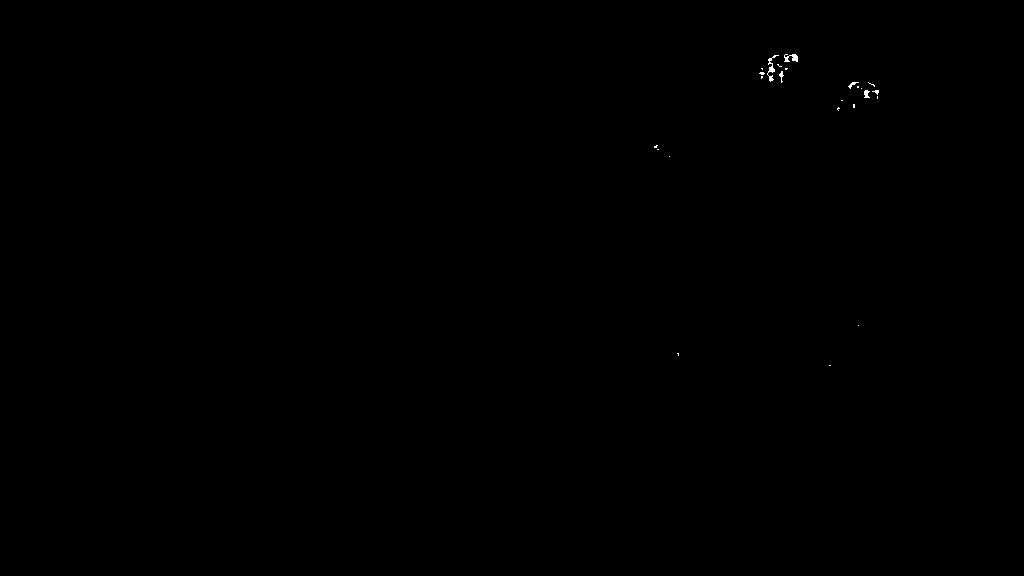}
        \caption*{}
      \end{subfigure}
      \hfill
      \begin{subfigure}[b]{0.20\linewidth}
        \centering
        \includegraphics[width=\linewidth]{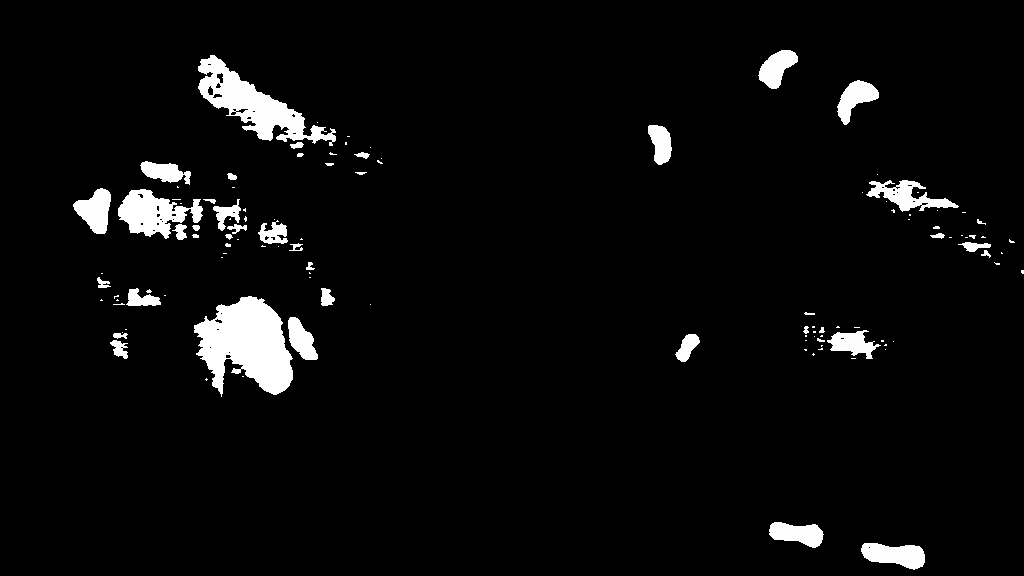}
        \caption*{}
      \end{subfigure}      
    \end{minipage}
  }
    \caption{
    \textbf{Comparison of segmentation results across five datasets}: polyp, leaf, eyes, chest, golf at 60 seconds. The figure highlights the performance consistency of QTT-SEG across diverse domains. All labels are white and background is black.
    }
    \label{fig:predictions_grid}
\end{figure}

\section{Introduction}
\label{introduction}

Pre-trained foundation models like SAM \citep{ravi2024sam} have revolutionized image segmentation by offering strong generalization across diverse domains. However, their performance often plateaus on specialized, domain-specific datasets where fine-tuning is necessary. Traditional adaptation methods typically require extensive manual tuning of hyperparameters and strategies \citep{hutter2015beyond, quinton2024navigating, sharma2019hyperparameter, ogundokun2022human}, making them resource-intensive and difficult to scale.

Quick-Tune \citep{arango2023quick} addresses this bottleneck through meta-learned predictors that guide efficient hyperparameter selection, previously demonstrated for image classification and extended by \citet{strangmann2024transfer} to the language domain. In this work, we extend Quick-Tune to image segmentation and show that it can effectively adapt SAM across a wide range of domains with minimal manual effort. Evaluated on 13 benchmark segmentation datasets, Quick-Tune consistently improves over SAM’s zero-shot performance and frequently outperforms AutoGluon — even under tight time budgets. These results underscore the potential of meta-learning to automate the adaptation of foundation models, enabling efficient and scalable performance optimization across diverse tasks and domains.

\section{Background and Related Work}
\paragraph{Quick-Tune} 
Quick-Tune \citep{arango2023quick} is a Bayesian Optimization (BO) framework for optimizing deep learning pipelines. It employs a probabilistic performance predictor $\hat{\ell}_\theta$ and a cost predictor $\hat{c}_w$ to select pipelines maximizing Multi-fidelity Expected Improvement while accounting for fine-tuning costs. Meta-learned parameters $\theta$ and $w$ guide exploration of hyperparameter and model configurations. Quick-Tune uses deep-kernel GPs \citep{wistuba2022supervising} and MLPs trained on historical data for performance and cost prediction. Initially applied to image classification, it has been extended to LLMs \citep{strangmann2024transfer} and released as an open-source tool \textit{Quick-Tune-Tool} \citep{rapant2024quick}. Here, we extend it to semantic segmentation.

\paragraph{Fine-tuning SAM}
Several methods adapt SAM \citep{ravi2024sam} using parameter-efficient tuning (e.g., SAM-PARSER \citep{10.1609/aaai.v38i5.28250}), generalization techniques (e.g., GSAM \citep{kato2024generalized}), and few-shot strategies \citep{xie2024masksam}. These enhance cross-domain performance with limited labels. Our approach automates SAM tuning via meta-learning and BO, reducing manual effort while maintaining strong results.

\paragraph{AutoML Systems} 
Tools like AutoKeras \citep{jin2019auto}, AutoPytorch \citep{zimmer2021auto}, and NePS \citep{mallik2023priorband} automate model and hyperparameter search. Others like AutoGluon MultiModal \citep{tang2024autogluon} and ZAP \citep{ozturk2022zero} focus on foundation model tuning. Unlike these, Quick-Tune leverages meta-learning of dataset meta-features and learning curves to guide search, enabling faster and more cost-efficient adaptation.

\section{Quick-Tune-Tool for Semantic Segmentation (QTT-SEG)}
Foundation model tuning for segmentation is difficult due to dataset diversity, resource constraints, and sensitivity to hyperparameters. QTT-SEG addresses this by efficiently tuning SAM \citep{ravi2024sam} through a structured BO pipeline (see \Cref{fig:flow}). It defines a rich hyperparameter space, trains performance predictors on learning curves, and evaluates robustness across datasets. We prompt SAM with bounding boxes extracted from ground truth masks, adding random perturbations to simulate noisy prompts.

\textbf{Search Space:}
We designed an extensive search space covering fine-tuning, optimization, and augmentation hyperparameters to maximize SAM’s performance on new datasets. The search space spans over 200 million configurations. Key groups include LoRA \citep{hu2022lora} application, data augmentation, optimizer, loss, learning rates, schedulers, and warm-up phases (see \Cref{tab:app-search_space}).

\textbf{Meta Training:}
We sampled $2\,000$ configuration–dataset pairs across all selected binary and multiclass segmentation datasets. Each sampled configuration was fine-tuned on its corresponding dataset for 10 epochs, while tracking performance and training cost. This resulted in a meta-dataset of configuration–performance and cost pairs, which were then used to pre-train predictors. 

\section{Experiments and Results}

\paragraph{Benchmark Setup}
We evaluate QTT-SEG on 8 binary and 5 multiclass semantic segmentation datasets that we manually curated (details in ~\Cref{appendix-datasets}).

\paragraph{Experimental Setup}
Each experiment uses 128 configurations, tuned under time budgets of 60, 120, and 180 seconds. We subsample 100 images and masks per dataset using five random seeds. Tuning is performed independently for each seed using QTT-SEG, reporting mean IoU and standard deviation. For each run, we exclude the target dataset’s learning curves from metadata, ensuring the predictors are trained only on other datasets, hence enabling generalization to unseen data.

For binary segmentation tasks, we compare QTT-SEG against both AutoGluon Multimodal \citep{tang2024autogluon} and SAM’s zero-shot performance, using consistent time budgets and random seeds. For multiclass tasks, only SAM zero-shot is used as a baseline, as AutoGluon’s support for multiclass segmentation is not clearly documented. We ran all experiments on a single NVIDIA GeForce RTX 2080 Ti GPU with 11 GB RAM.

\paragraph{Results}
\Cref{fig:avg_performance} compares average performance of QTT-SEG on all datasets with baselines. Table \Cref{tab:segmentation_results_sorted} shows QTT-SEG’s performance versus AutoGluon and zero-shot SAM on binary datasets and Table \Cref{tab:segmentation_results_multi_sorted} covers multiclass datasets. \begin{enumerate*}[label=(\roman*)]
\item \textbf{QTT-SEG consistently outperforms zero-shot baselines across all datasets.}  
On average, QTT-SEG achieves a 76.47\% improvement in IoU over zero-shot baselines across all datasets at 180 seconds.
\item \textbf{QTT-SEG is a compute-efficient alternative.}  
QTT-SEG reaches 97.3\% of its final accuracy within 60 seconds and maintains better accuracy than AutoGluon on 6 out of 8 binary classification datasets at the 180-second mark. This highlights QTT-SEG’s efficiency compared to AutoGluon’s incremental but slower convergence, likely due to QTT-SEG’s meta-trained predictors enabling faster convergence to superior configurations, while AutoGluon relies on a more exhaustive search.
\item \textbf{QTT-SEG is robust to local optima.}  
The relative gain of 1.97\% between 120 and 180 seconds, compared to a smaller 0.95\% gain between 60 and 120 seconds, (avg. on binary) suggests that QTT-SEG continues to improve beyond early plateaus. While relative gains alone do not prove escaping local optima, this pattern indicates effective refinement of configurations and is further supported by superior performance on 6 out of 8 datasets. 
\item \textbf{QTT-SEG demonstrates stronger domain generalization.}  
Although AutoGluon performs better on datasets like `chest' and `eyes', QTT-SEG outperforms it on the others. \Cref{fig:predictions_grid} shows representative results where each framework excels. With a higher average improvement over zero-shot (67.16\% vs. 55.72\% at 180 seconds), QTT-SEG appears more robust to domain shifts overall. See \Cref{appendix-all-masks} for sample predictions on all datasets. 
\item \textbf{QTT-SEG scales effectively to multiclass datasets.} 
On multiclass segmentation tasks, QTT-SEG surpasses zero-shot baselines by an average of 85.28\% at 180 seconds.
\item \textbf{QTT-SEG demonstrates stability across folds.} 
QTT-SEG proves to be a highly reliable tuning framework, as indicated by its low standard deviation across folds within the same dataset.
\end{enumerate*}

\paragraph{Conclusion}
QTT-SEG is an efficient AutoML framework that consistently outperforms zero-shot baselines and converges faster than AutoGluon across diverse segmentation tasks. Its meta-trained predictors enable rapid tuning and robust escape from local optima, leading to superior accuracy and stability. While AutoGluon excels on certain modalities, QTT-SEG shows greater overall robustness to domain shifts. Future work would focus on expanding the search space and incorporating richer domain knowledge to further improve performance.

\newpage
\begin{acknowledgements}
L.P. acknowledges funding by the Deutsche Forschungsgemeinschaft (DFG, German Research Foundation) under SFB 1597 (SmallData), grant number 499552394. 
F.H. acknowledges the financial support of the Hector Foundation. 
\end{acknowledgements}

\bibliography{automl-conf/references}

\begin{thebibliography}{}

\bibitem[Arango et~al., 2023]{arango2023quick}
Arango, S.~P., Ferreira, F., Kadra, A., Hutter, F., and Grabocka, J. (2023).
\newblock Quick-tune: Quickly learning which pretrained model to finetune and how.
\newblock {\em arXiv preprint arXiv:2306.03828}.

\bibitem[Degerli et~al., 2022]{degerli2022osegnet}
Degerli, A., Kiranyaz, S., Chowdhury, M.~E., and Gabbouj, M. (2022).
\newblock Osegnet: Operational segmentation network for covid-19 detection using chest x-ray images.
\newblock In {\em 2022 IEEE International Conference on Image Processing (ICIP)}, pages 2306--2310. IEEE.

\bibitem[Hu et~al., 2022]{hu2022lora}
Hu, E.~J., Shen, Y., Wallis, P., Allen-Zhu, Z., Li, Y., Wang, S., Wang, L., Chen, W., et~al. (2022).
\newblock Lora: Low-rank adaptation of large language models.
\newblock {\em ICLR}, 1(2):3.

\bibitem[Hutter et~al., 2015]{hutter2015beyond}
Hutter, F., L{\"u}cke, J., and Schmidt-Thieme, L. (2015).
\newblock Beyond manual tuning of hyperparameters.
\newblock {\em KI-K{\"u}nstliche Intelligenz}, 29:329--337.

\bibitem[Jin et~al., 2019]{jin2019auto}
Jin, H., Song, Q., and Hu, X. (2019).
\newblock Auto-keras: An efficient neural architecture search system.
\newblock In {\em Proceedings of the 25th ACM SIGKDD international conference on knowledge discovery \& data mining}, pages 1946--1956.

\bibitem[Kato et~al., 2024]{kato2024generalized}
Kato, S., Mitsuoka, H., and Hotta, K. (2024).
\newblock Generalized sam: Efficient fine-tuning of sam for variable input image sizes.
\newblock {\em arXiv preprint arXiv:2408.12406}.

\bibitem[Kingma and Ba, 2014]{kingma2014adam}
Kingma, D.~P. and Ba, J. (2014).
\newblock Adam: A method for stochastic optimization.
\newblock {\em arXiv preprint arXiv:1412.6980}.

\bibitem[Loshchilov and Hutter, 2016]{loshchilov2016sgdr}
Loshchilov, I. and Hutter, F. (2016).
\newblock Sgdr: Stochastic gradient descent with warm restarts.
\newblock {\em arXiv preprint arXiv:1608.03983}.

\bibitem[Mallik et~al., 2023]{mallik2023priorband}
Mallik, N., Bergman, E., Hvarfner, C., Stoll, D., Janowski, M., Lindauer, M., Nardi, L., and Hutter, F. (2023).
\newblock Priorband: Practical hyperparameter optimization in the age of deep learning.
\newblock In {\em Thirty-seventh Conference on Neural Information Processing Systems (NeurIPS 2023)}.

\bibitem[Ogundokun et~al., 2022]{ogundokun2022human}
Ogundokun, R.~O., Maskeli{\=u}nas, R., and Dama{\v{s}}evi{\v{c}}ius, R. (2022).
\newblock Human posture detection using image augmentation and hyperparameter-optimized transfer learning algorithms.
\newblock {\em Applied Sciences}, 12(19):10156.

\bibitem[{\"O}zt{\"u}rk et~al., 2022]{ozturk2022zero}
{\"O}zt{\"u}rk, E., Ferreira, F., Jomaa, H., Schmidt-Thieme, L., Grabocka, J., and Hutter, F. (2022).
\newblock Zero-shot automl with pretrained models.
\newblock In {\em International Conference on Machine Learning}, pages 17138--17155. PMLR.

\bibitem[Paszke, 2019]{paszke2019pytorch}
Paszke, A. (2019).
\newblock Pytorch: An imperative style, high-performance deep learning library.
\newblock {\em arXiv preprint arXiv:1912.01703}.

\bibitem[Peng et~al., 2024]{10.1609/aaai.v38i5.28250}
Peng, Z., Xu, Z., Zeng, Z., Yang, X., and Shen, W. (2024).
\newblock Sam-parser: fine-tuning sam efficiently by parameter space reconstruction.
\newblock In {\em Proceedings of the Thirty-Eighth AAAI Conference on Artificial Intelligence and Thirty-Sixth Conference on Innovative Applications of Artificial Intelligence and Fourteenth Symposium on Educational Advances in Artificial Intelligence}, AAAI'24/IAAI'24/EAAI'24. AAAI Press.

\bibitem[Quinton et~al., 2024]{quinton2024navigating}
Quinton, F., Presles, B., Leclerc, S., Nodari, G., Lopez, O., Chevallier, O., Pellegrinelli, J., Vrigneaud, J.-M., Popoff, R., Meriaudeau, F., et~al. (2024).
\newblock Navigating the nuances: comparative analysis and hyperparameter optimisation of neural architectures on contrast-enhanced mri for liver and liver tumour segmentation.
\newblock {\em Scientific Reports}, 14(1):3522.

\bibitem[Rapant et~al., 2024]{rapant2024quick}
Rapant, I., Purucker, L., Ferreira, F., Arango, S.~P., Kadra, A., Grabocka, J., and Hutter, F. (2024).
\newblock Quick-tune-tool: A practical tool and its user guide for automatically finetuning pretrained models.
\newblock In {\em AutoML Conference 2024 (Workshop Track)}.

\bibitem[Ravi et~al., 2024]{ravi2024sam}
Ravi, N., Gabeur, V., Hu, Y.-T., Hu, R., Ryali, C., Ma, T., Khedr, H., R{\"a}dle, R., Rolland, C., Gustafson, L., et~al. (2024).
\newblock Sam 2: Segment anything in images and videos.
\newblock {\em arXiv preprint arXiv:2408.00714}.

\bibitem[Sharma et~al., 2019]{sharma2019hyperparameter}
Sharma, A., van Rijn, J.~N., Hutter, F., and M{\"u}ller, A. (2019).
\newblock Hyperparameter importance for image classification by residual neural networks.
\newblock In {\em Discovery Science: 22nd International Conference, DS 2019, Split, Croatia, October 28--30, 2019, Proceedings 22}, pages 112--126. Springer.

\bibitem[Smith and Topin, 2019]{smith2019super}
Smith, L.~N. and Topin, N. (2019).
\newblock Super-convergence: Very fast training of neural networks using large learning rates.
\newblock In {\em Artificial intelligence and machine learning for multi-domain operations applications}, volume 11006, pages 369--386. SPIE.

\bibitem[Strangmann et~al., 2024]{strangmann2024transfer}
Strangmann, T., Purucker, L., Franke, J.~K., Rapant, I., Ferreira, F., and Hutter, F. (2024).
\newblock Transfer learning for finetuning large language models.
\newblock In {\em Adaptive Foundation Models: Evolving AI for Personalized and Efficient Learning}.

\bibitem[Sudre et~al., 2017]{sudre2017generalised}
Sudre, C.~H., Li, W., Vercauteren, T., Ourselin, S., and Jorge~Cardoso, M. (2017).
\newblock Generalised dice overlap as a deep learning loss function for highly unbalanced segmentations.
\newblock In {\em Deep Learning in Medical Image Analysis and Multimodal Learning for Clinical Decision Support: Third International Workshop, DLMIA 2017, and 7th International Workshop, ML-CDS 2017, Held in Conjunction with MICCAI 2017, Qu{\'e}bec City, QC, Canada, September 14, Proceedings 3}, pages 240--248. Springer.

\bibitem[Tang et~al., 2024]{tang2024autogluon}
Tang, Z., Fang, H., Zhou, S., Yang, T., Zhong, Z., Hu, T., Kirchhoff, K., and Karypis, G. (2024).
\newblock Autogluon-multimodal (automm): Supercharging multimodal automl with foundation models.
\newblock {\em arXiv preprint arXiv:2404.16233}.

\bibitem[Wagner, 2023]{franz_wagner_2023}
Wagner, F. (2023).
\newblock Fiber segmentation dataset.

\bibitem[Wistuba et~al., 2022]{wistuba2022supervising}
Wistuba, M., Kadra, A., and Grabocka, J. (2022).
\newblock Supervising the multi-fidelity race of hyperparameter configurations.
\newblock {\em Advances in Neural Information Processing Systems}, 35:13470--13484.

\bibitem[Xie et~al., 2024]{xie2024masksam}
Xie, B., Tang, H., Duan, B., Cai, D., and Yan, Y. (2024).
\newblock Masksam: Towards auto-prompt sam with mask classification for medical image segmentation.
\newblock {\em arXiv preprint arXiv:2403.14103}.

\bibitem[Zimmer et~al., 2021]{zimmer2021auto}
Zimmer, L., Lindauer, M., and Hutter, F. (2021).
\newblock Auto-pytorch: Multi-fidelity metalearning for efficient and robust autodl.
\newblock {\em IEEE transactions on pattern analysis and machine intelligence}, 43(9):3079--3090.

\end{thebibliography}

\newpage
\section{[Optional] Submission Checklist}

\begin{enumerate}
\item For all authors\dots
  \begin{enumerate}
  \item Do the main claims made in the abstract and introduction accurately
    reflect the paper's contributions and scope?
    \answerYes{}
  \item Did you describe the limitations of your work?
    \answerYes{}
  \item Did you discuss any potential negative societal impacts of your work?
    \answerNA{}
  \item Did you read the ethics review guidelines and ensure that your paper
    conforms to them? (see \url{https://2022.automl.cc/ethics-accessibility/})
    \answerYes{}
  \end{enumerate}
\item If you ran experiments\dots
  \begin{enumerate}
  \item Did you use the same evaluation protocol for all methods being compared (e.g.,
    same benchmarks, data (sub)sets, available resources, etc.)?
    \answerYes{}
  \item Did you specify all the necessary details of your evaluation (e.g., data splits,
    pre-processing, search spaces, hyperparameter tuning details and results, etc.)?
    \answerYes{}
  \item Did you repeat your experiments (e.g., across multiple random seeds or
    splits) to account for the impact of randomness in your methods or data?
    \answerYes{}
  \item Did you report the uncertainty of your results (e.g., the standard error
    across random seeds or splits)?
    \answerYes{}
  \item Did you report the statistical significance of your results?
    \answerNA{}
  \item Did you use enough repetitions, datasets, and/or benchmarks to support
    your claims?
    \answerYes{}
  \item Did you compare performance over time and describe how you selected the
    maximum runtime?
    \answerYes{}
  \item Did you include the total amount of compute and the type of resources
    used (e.g., type of \textsc{gpu}s, internal cluster, or cloud provider)?
    \answerYes{}
  \item Did you run ablation studies to assess the impact of different
    components of your approach?
    \answerYes{}
  \end{enumerate}
\item With respect to the code used to obtain your results\dots
  \begin{enumerate}
\item Did you include the code, data, and instructions needed to reproduce the
    main experimental results, including all dependencies (e.g.,
    \texttt{requirements.txt} with explicit versions), random seeds, an instructive
    \texttt{README} with installation instructions, and execution commands
    (either in the supplemental material or as a \textsc{url})?
    \answerYes{}
  \item Did you include a minimal example to replicate results on a small subset
    of the experiments or on toy data?
    \answerYes{}
  \item Did you ensure sufficient code quality and documentation so that someone else
    can execute and understand your code?
    \answerYes{}
  \item Did you include the raw results of running your experiments with the given
    code, data, and instructions?
    \answerYes{}
  \item Did you include the code, additional data, and instructions needed to generate
    the figures and tables in your paper based on the raw results?
    \answerYes{}
  \end{enumerate}
\item If you used existing assets (e.g., code, data, models)\dots
  \begin{enumerate}
  \item Did you citep the creators of used assets?
    \answerYes{}
  \item Did you discuss whether and how consent was obtained from people whose
    data you're using/curating if the license requires it?
    \answerNA{}
  \item Did you discuss whether the data you are using/curating contains
    personally identifiable information or offensive content?
    \answerNA{}
  \end{enumerate}
\item If you created/released new assets (e.g., code, data, models)\dots
  \begin{enumerate}
    \item Did you mention the license of the new assets (e.g., as part of your
    code submission)?
    \answerYes{}
    \item Did you include the new assets either in the supplemental material or as
    a \textsc{url} (to, e.g., GitHub or Hugging Face)?
    \answerYes{}
  \end{enumerate}
\item If you used crowdsourcing or conducted research with human subjects\dots
  \begin{enumerate}
  \item Did you include the full text of instructions given to participants and
    screenshots, if applicable?
    \answerNA{}
  \item Did you describe any potential participant risks, with links to
    institutional review board (\textsc{irb}) approvals, if applicable?
    \answerNA{}
  \item Did you include the estimated hourly wage paid to participants and the
    total amount spent on participant compensation?
    \answerNA{}
  \end{enumerate}
\item If you included theoretical results\dots
  \begin{enumerate}
  \item Did you state the full set of assumptions of all theoretical results?
    \answerNA{}
  \item Did you include complete proofs of all theoretical results?
    \answerNA{}
  \end{enumerate}
\end{enumerate}

\newpage
\appendix

\section{Search-Space}
\label{appendix-search-space}

The QTT-SEG search space is designed to enable flexible and effective adaptation of large segmentation models across diverse domains and computational budgets. It includes a range of architectural and training hyperparameters, with a focus on efficiency and robustness. Core components of the space involve the optional use of LoRA (Low-Rank Adaptation), applied to the attention and MLP layers of the image encoder, with tunable settings for activation, rank, and dropout. Optimization is performed using the AdamW optimizer \citep{kingma2014adam}, with a finely grained range of learning rates spanning from 1e-5 to 7e-3, sampled in a rough-logarithmic progression to allow fine control in both low and high learning rate regimes. Data augmentation strategies such as horizontal and vertical flips, and random rotations are included as binary choices. The loss function combines Binary Cross-Entropy (BCE) with Dice loss \citep{sudre2017generalised} to handle class imbalance common in segmentation tasks. A variety of learning rate schedulers are supported: Cosine \citep{loshchilov2016sgdr}, OneCycle \citep{smith2019super}, Plateau, Step and Polynomial \citep{paszke2019pytorch}, each with its own tunable parameters to better adapt to dataset and time-budget characteristics. This rich and modular search space, detailed in Table~\Cref{tab:app-search_space}, allows QTT-SEG to discover effective configurations in a time-constrained tuning setting.

\begin{table*}[t]
\caption{\textbf{QTT-SEG Search Space:} Key hyperparameters for segmentation tuning.}
\label{tab:app-search_space}
\centering
\begin{small}
\begin{tabularx}{\textwidth}{lX}
\toprule
\textbf{Hyper-Parameters} & \textbf{Choices} \\
\midrule
LoRA Application & Image Encoder Attention \& MLP Layers \\
LoRA Enabled & 0, 1 \\
LoRA Rank & 4, 8, 16 \\
LoRA Dropout & 0.0, 0.1 \\
Optimizer & AdamW \\
Weight Decay & 0.0, 1e-5, 5e-5, 1e-4 \\
Learning Rate & 1e-5, 1.2e-5, 1.5e-5, 2e-5, 2.5e-5, 3.5e-5, 5e-5, 6e-5, 6.5e-5, 0.0001, 0.00012, 0.00018, 0.00025, 0.00032, 0.0004, 0.00048, 0.0005, 0.00055, 0.0008, 0.001, 0.0015, 0.002, 0.003, 0.004, 0.005, 0.006, 0.007 \\
Loss Function & BCE + Dice \\
Data Augmentation & Horizontal Flip (0,1), Vertical Flip (0,1), Random Rotate (0,1) \\
Learning Rate Schedulers & Cosine, OneCycle, Plateau, Cosine\_Warm, Step, Poly \\
\midrule
Plateau Parameters: & \begin{tabular}[t]{@{}l@{}}
\hspace{1em}factor: 0.1, 0.5, 0.8 \\
\hspace{1em}patience: 0, 1, 2
\end{tabular} \\
Cosine\_Warm Parameters: & \begin{tabular}[t]{@{}l@{}}
\hspace{1em}$T_0$: 2, 3, 5 \\
\hspace{1em}$T_{\text{mult}}$: 1, 2
\end{tabular} \\
OneCycle Parameters: & \begin{tabular}[t]{@{}l@{}}
\hspace{1em}pct\_start: 0.030–0.100 (step 0.005) \\
\hspace{1em}div\_factor: 10–100 \\
\hspace{1em}final\_div\_factor: 10–1000
\end{tabular} \\
Step Scheduler Parameters: & \hspace{1em}step\_size: 3, 5 \\
Polynomial Scheduler Parameters: & \hspace{1em}power: 0.5, 0.9, 1.0 \\
\bottomrule
\end{tabularx}
\end{small}
\vskip -0.1in
\end{table*}

\section{Datasets}
\label{appendix-datasets}

We present a brief overview and sources of each dataset used to benchmark QTT-SEG's performance. Our selection prioritizes diversity across domains and real-world relevance in semantic segmentation tasks, while ensuring all datasets are openly accessible from reputable platforms such as Kaggle, Hugging Face, and DatasetNinja.

\paragraph{Binary} We tested on eight binary classification datasets:
\begin{itemize}
    
    \item \textbf{polyp}: This dataset comprises 612 colonoscopy frames extracted from 29 video sequences, each annotated with a binary mask highlighting polyps. It is designed for semantic segmentation tasks in medical imaging, particularly for polyp detection in colonoscopy videos. The dataset is valuable for training and evaluating models in early colorectal cancer detection. Source: \href{https://www.kaggle.com/datasets/balraj98/cvcclinicdb}{CVC-ClinicDB on Kaggle.}

    \item \textbf{lesion}: This dataset comprises 900 dermoscopic images of skin lesions, each paired with a corresponding binary segmentation mask. The images are provided in JPEG format, while the masks are in PNG format, with pixel values of 0 for background and 255 for the lesion area. This dataset was used in the ISIC 2016 Challenge for automated skin lesion segmentation. Source:  \href{https://www.kaggle.com/datasets/santiagodelrey/isic-2016-task-1-training-data}{ISIC 2016 Task 1 Training Data on Kaggle.}

    \item \textbf{leaf}: This dataset comprises 588 images of diseased leaves, each paired with a corresponding segmentation mask. The data collection is based on the PlantDoc images, making it suitable for training and evaluating semantic segmentation models in plant disease detection tasks. Source: \href{https://www.kaggle.com/datasets/fakhrealam9537/leaf-disease-segmentation-dataset}{Leaf Disease Segmentation Dataset on Kaggle.}

    \item \textbf{covid} \citep{degerli2022osegnet}: This extensive dataset comprises over 121,000 chest X-ray images, including 9,258 annotated COVID-19 cases, sourced from Qatar University, Tampere University, and Hamad Medical Corporation. It supports both classification and segmentation tasks, featuring ground-truth masks for lesion areas. The dataset is valuable for training and evaluating deep learning models in medical image analysis. Source: \href{https://www.kaggle.com/datasets/aysendegerli/qatacov19-dataset}{QaTa-COV19 on Kaggle.}

    \item \textbf{eyes}: This dataset comprises 962 retinal images annotated for instance segmentation, focusing on microvascular structures. It includes 4,277 labeled vessels and 313 unlabeled images, with a 2:1 train-test split. The dataset is suitable for training and evaluating models in medical image segmentation tasks. Source: \href{https://datasetninja.com/eyes-microcirculation}{Eyes Microcirculation on Dataset Ninja.}

    \item \textbf{fiber} \citep{franz_wagner_2023}: This dataset consists of 3 spatially disjoint volumes, each with dimensions 20 × 512 × 512 voxels (voxel size: 4~\(\mu\)m). It is designed for fiber segmentation tasks in CT scans of concrete, providing a valuable resource for training and evaluating segmentation models in material science applications. Source: \href{https://www.kaggle.com/datasets/franzwagner/pe-fibers?select=fibers_original}{PE-Fibers on Kaggle.}

    \item \textbf{cardiac}: This dataset comprises X-ray images of cardiac catheterization procedures, each with dimensions of 512x512 pixels in PNG format. The images are designed for training and evaluating models in medical image analysis tasks, particularly in the context of cardiovascular imaging. Source: \href{https://www.kaggle.com/datasets/c7934597/cardiac-catheterization}{Cardiac Catheterization X-Ray PNG 512x512 on Kaggle.}

    \item \textbf{chest}: This dataset contains chest X-ray images with corresponding lung masks, designed for lung segmentation tasks. It includes both original and pre-processed grayscale images, along with binary masks for the left and right lungs. The dataset is suitable for training and evaluating medical image segmentation models. Source: \href{https://www.kaggle.com/datasets/iamtapendu/chest-x-ray-lungs-segmentation}{Chest X-ray Lung Segmentation on Kaggle.}

\end{itemize}

\paragraph{Multi-class} We tested on five multiclass datasets:

\begin{itemize}

    \item \textbf{US}: This dataset comprises 617 real abdominal ultrasound scans, each annotated with manual segmentations of several abdominal organs, including the liver, kidneys, gallbladder, and spleen. The dataset is designed for training and evaluating models in medical image segmentation tasks, particularly in the context of abdominal ultrasound imaging. Source: \href{https://www.kaggle.com/datasets/ignaciorlando/ussimandsegm}{USSimAndSegm on Kaggle.}

    \item \textbf{human\_parsing}: This dataset contains 17,706 images with corresponding segmentation masks, sourced from the ATR dataset. It is designed for human parsing tasks, providing detailed pixel-level annotations of clothing and body parts. The dataset is useful for training and evaluating human segmentation models. Source: \href{https://huggingface.co/datasets/mattmdjaga/human_parsing_dataset}{Human Parsing Dataset on Hugging Face.}

    \item \textbf{golf}: This dataset comprises 1,123 RGB orthophotos collected from 107 Danish golf courses during the spring season. The images are annotated for instance segmentation tasks, focusing on classes such as greens, fairways, tees, bunkers, and water hazards. An additional 108 images are provided for testing, with manual masking of non-course areas to facilitate course rating computations. The dataset is valuable for training and evaluating models in sports and geospatial analysis. Source: \href{https://www.kaggle.com/datasets/jacotaco/danish-golf-courses-orthophotos}{Danish Golf Courses Orthophotos on Kaggle.}

    \item \textbf{terrain}: This dataset comprises 1,000 synthetic images generated using the Unity engine, each annotated with pixel-level semantic labels. It is designed for training and evaluating semantic segmentation models in mobile robotics applications. The dataset includes various terrain types such as grass, dirt, and pavement, providing a diverse set of scenes for model training. Source: \href{https://www.kaggle.com/datasets/sadhoss/vale-semantic-terrain-segmentation}{Vale Semantic Terrain Segmentation on Kaggle.}
    
    \item \textbf{cholec}: This dataset is a semantic segmentation benchmark derived from the Cholec80 dataset. It consists of 8,080 annotated frames from 17 laparoscopic cholecystectomy videos, labeled for 13 different classes including surgical instruments and anatomical structures. The dataset is useful for evaluating real-time segmentation models in surgical scenes. Source: \href{https://www.kaggle.com/datasets/newslab/cholecseg8k}{CholecSeg8k on Kaggle.}

\end{itemize}

\section{Sample Prediction masks for all datasets}
\label{appendix-all-masks}
We present sample prediction masks for all datasets by QTT-SEG and compare with zero-shot baselines.

\begin{figure}[ht]
\setlength{\belowcaptionskip}{-5pt}
  \centering
  \resizebox{0.8\linewidth}{!}{ %
    \begin{minipage}{\textwidth} %

    \begin{subfigure}[b]{0.20\linewidth}
      \centering
      \makebox[0pt]{\textbf{Input}}
    \end{subfigure}
    \hfill
    \begin{subfigure}[b]{0.20\linewidth}
      \centering
      \makebox[0pt]{\textbf{Ground Truth}}
    \end{subfigure}
    \hfill
    \begin{subfigure}[b]{0.20\linewidth}
      \centering
      \makebox[0pt]{\textbf{SAM-Zero-Shot}}
    \end{subfigure}
    \hfill
    \begin{subfigure}[b]{0.20\linewidth}
      \centering
      \makebox[0pt]{\textbf{QTT-SEG}}
    \end{subfigure}
    
    \vspace{1mm}

    \centering
      \begin{subfigure}[b]{0.20\linewidth}
        \centering
        \includegraphics[width=\linewidth]{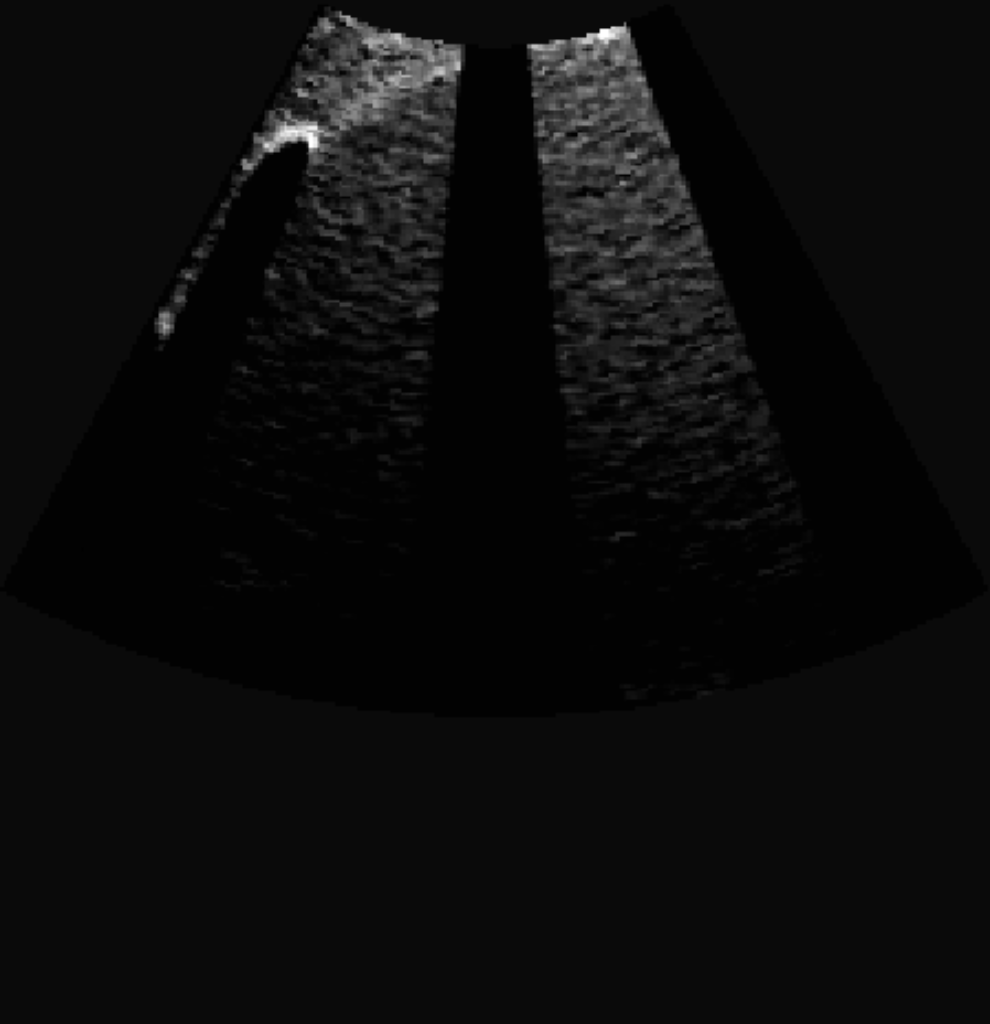}
        \caption*{}
      \end{subfigure}
      \hfill
      \begin{subfigure}[b]{0.20\linewidth}
        \centering
        \includegraphics[width=\linewidth]{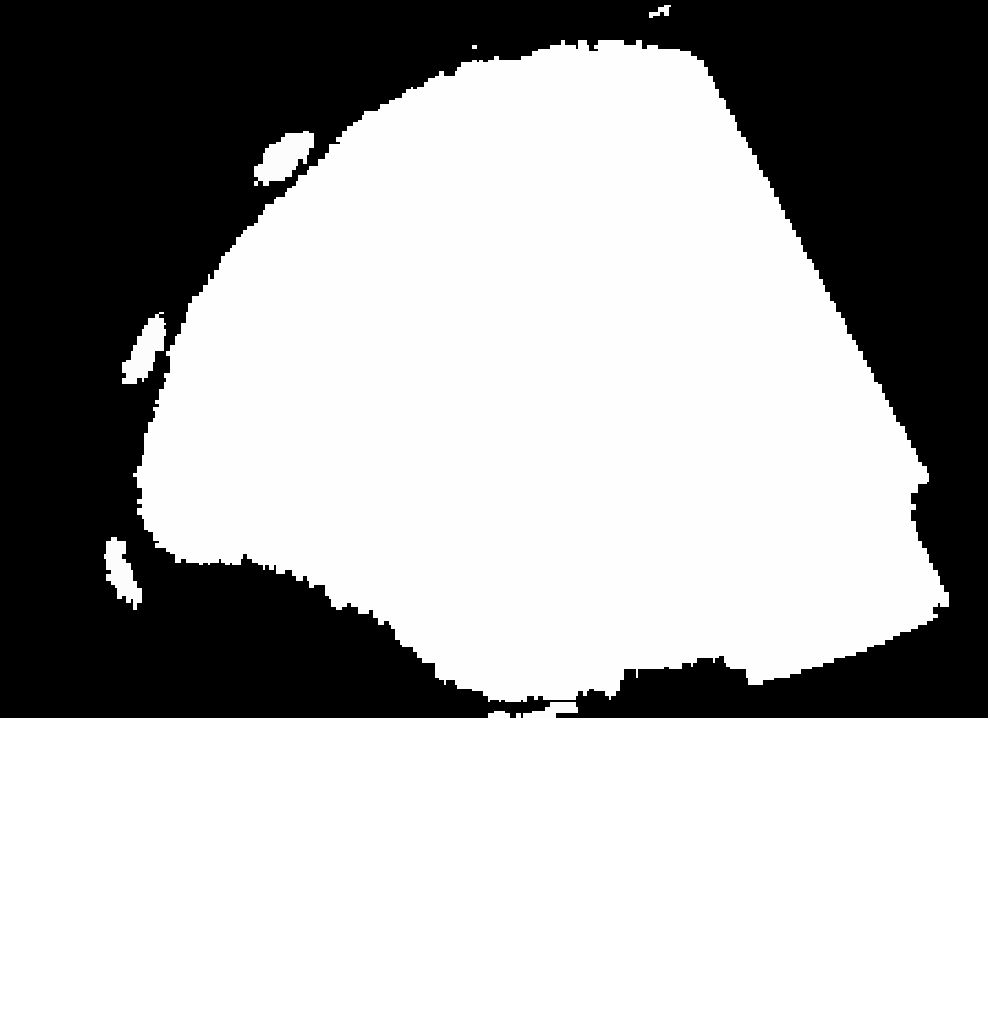}
        \caption*{}
      \end{subfigure}
      \hfill
      \begin{subfigure}[b]{0.20\linewidth}
        \centering
        \includegraphics[width=\linewidth]{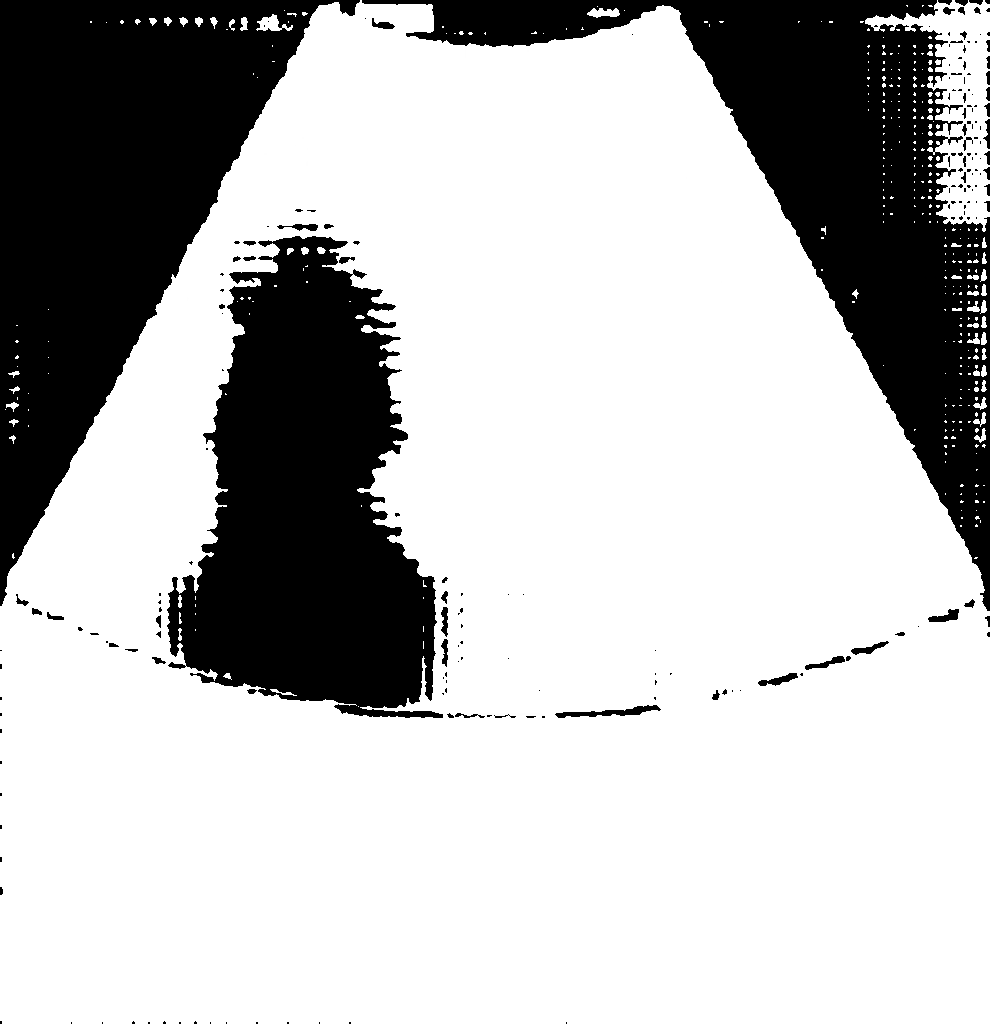}
        \caption*{}
      \end{subfigure}
      \hfill
      \begin{subfigure}[b]{0.20\linewidth}
        \centering
        \includegraphics[width=\linewidth]{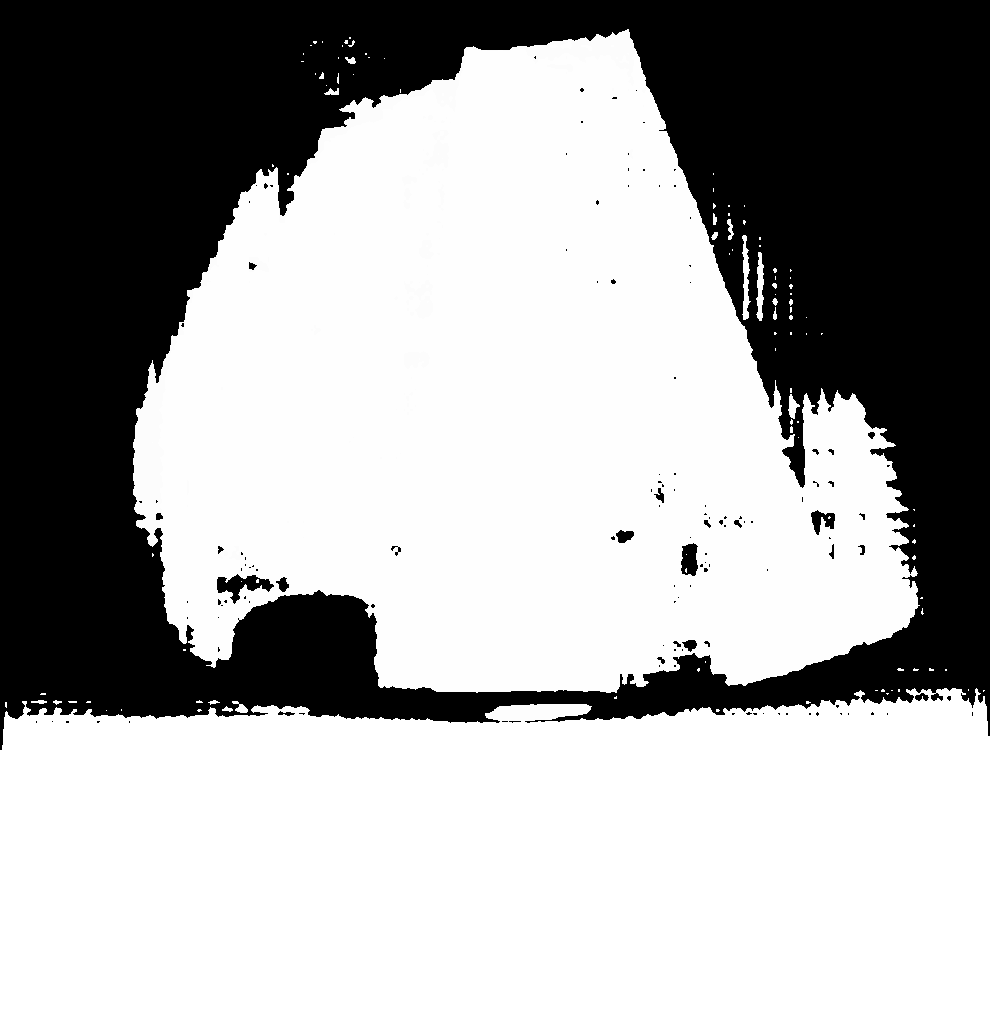}
        \caption*{}
      \end{subfigure}   

    \vspace{-4mm}
      \begin{subfigure}[b]{0.20\linewidth}
        \centering
        \includegraphics[width=\linewidth]{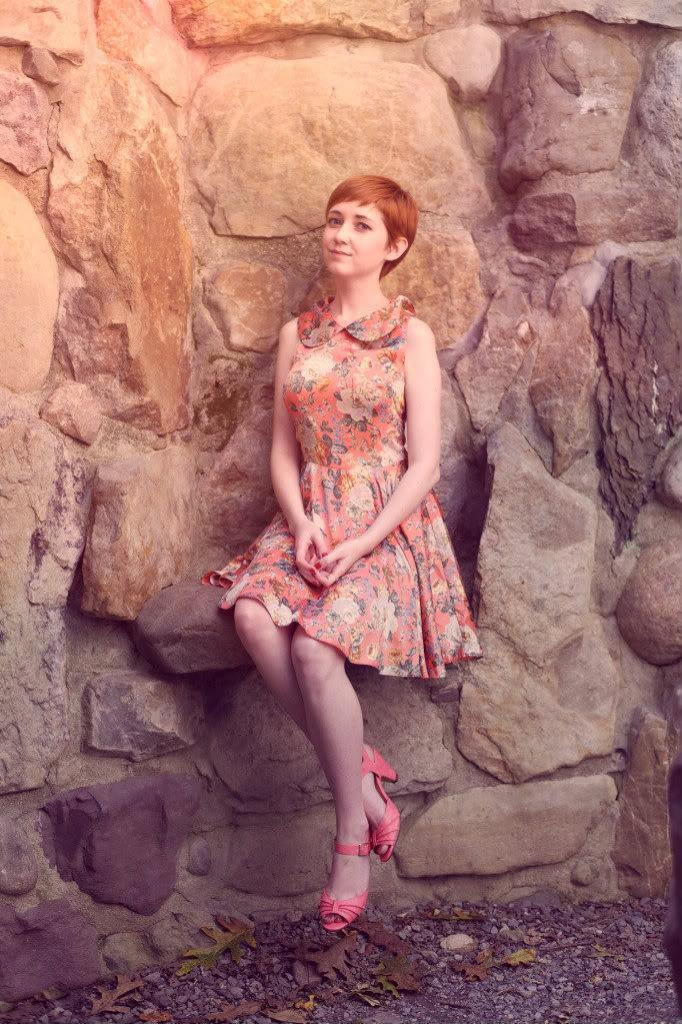}
        \caption*{}
      \end{subfigure}
      \hfill
      \begin{subfigure}[b]{0.20\linewidth}
        \centering
        \includegraphics[width=\linewidth]{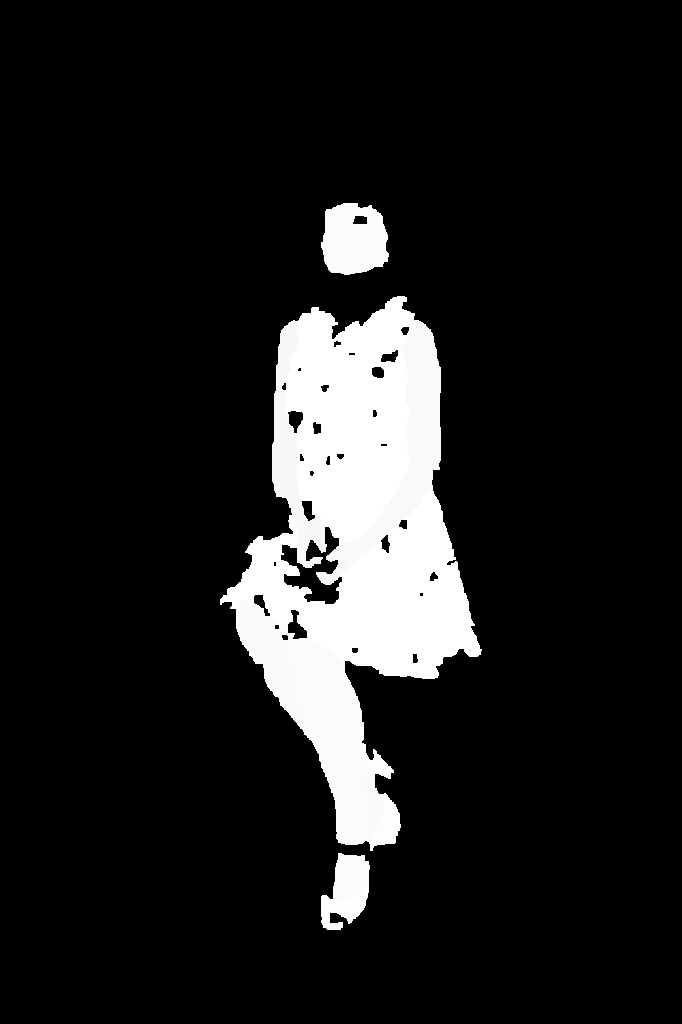}
        \caption*{}
      \end{subfigure}
      \hfill
      \begin{subfigure}[b]{0.20\linewidth}
        \centering
        \includegraphics[width=\linewidth]{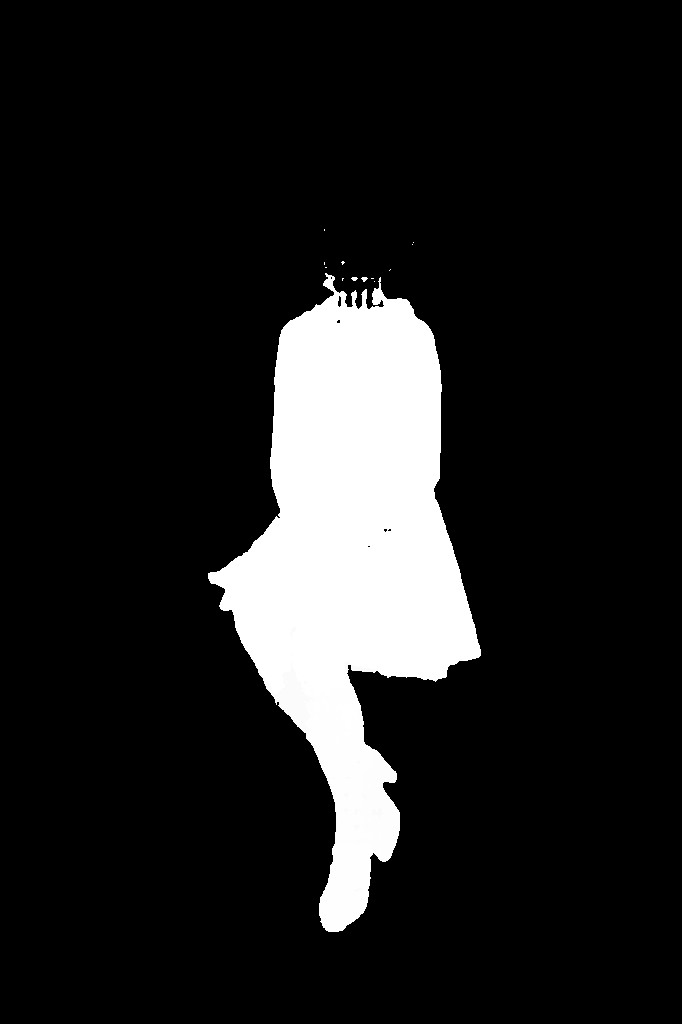}
        \caption*{}
      \end{subfigure}
      \hfill
      \begin{subfigure}[b]{0.20\linewidth}
        \centering
        \includegraphics[width=\linewidth]{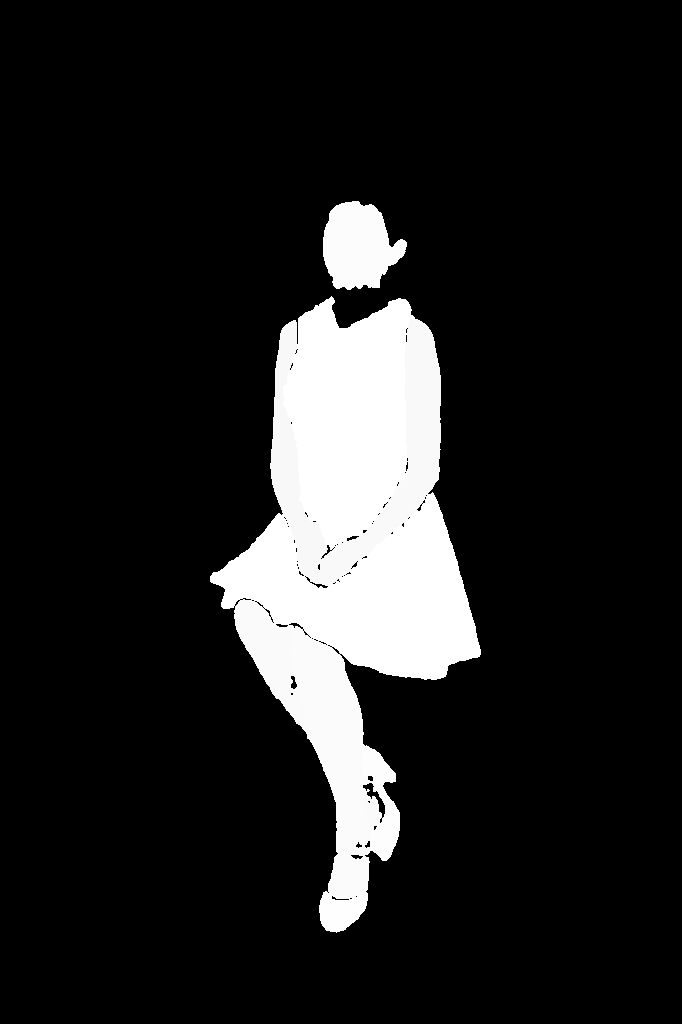}
        \caption*{}
      \end{subfigure} 

    \vspace{-4mm}
      \begin{subfigure}[b]{0.20\linewidth}
        \centering
        \includegraphics[width=\linewidth]{images/preds/golf/image_1.png}
        \caption*{}
      \end{subfigure}
      \hfill
      \begin{subfigure}[b]{0.20\linewidth}
        \centering
        \includegraphics[width=\linewidth]{images/preds/golf/gt_mask_1.png}
        \caption*{}
      \end{subfigure}
      \hfill
      \begin{subfigure}[b]{0.20\linewidth}
        \centering
        \includegraphics[width=\linewidth]{images/preds/golf/prd_mask_1_zero.png}
        \caption*{}
      \end{subfigure}
      \hfill
      \begin{subfigure}[b]{0.20\linewidth}
        \centering
        \includegraphics[width=\linewidth]{images/preds/golf/prd_mask_1.png}
        \caption*{}
      \end{subfigure} 

    \vspace{-4mm}
      \begin{subfigure}[b]{0.20\linewidth}
        \centering
        \includegraphics[width=\linewidth]{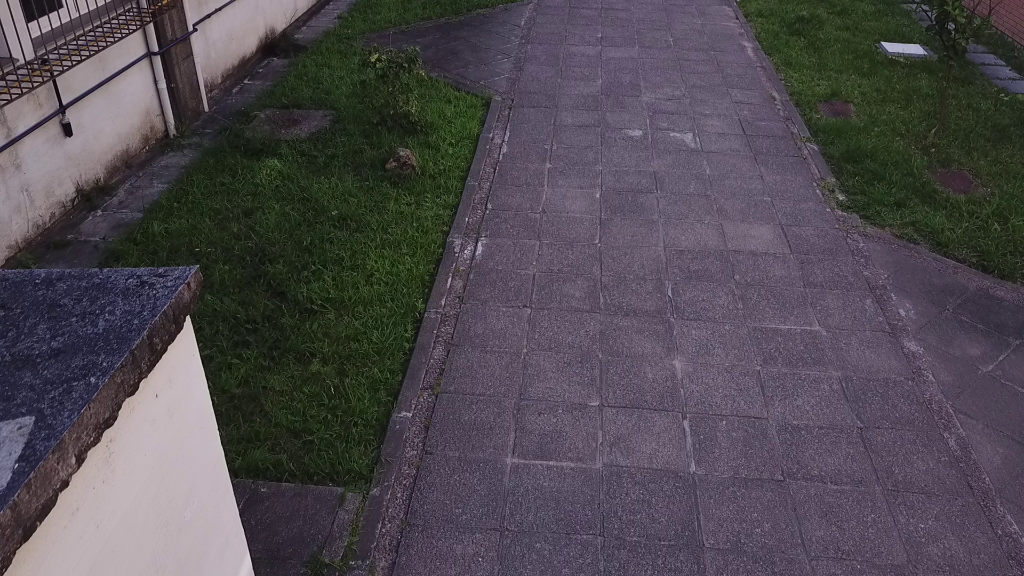}
        \caption*{}
      \end{subfigure}
      \hfill
      \begin{subfigure}[b]{0.20\linewidth}
        \centering
        \includegraphics[width=\linewidth]{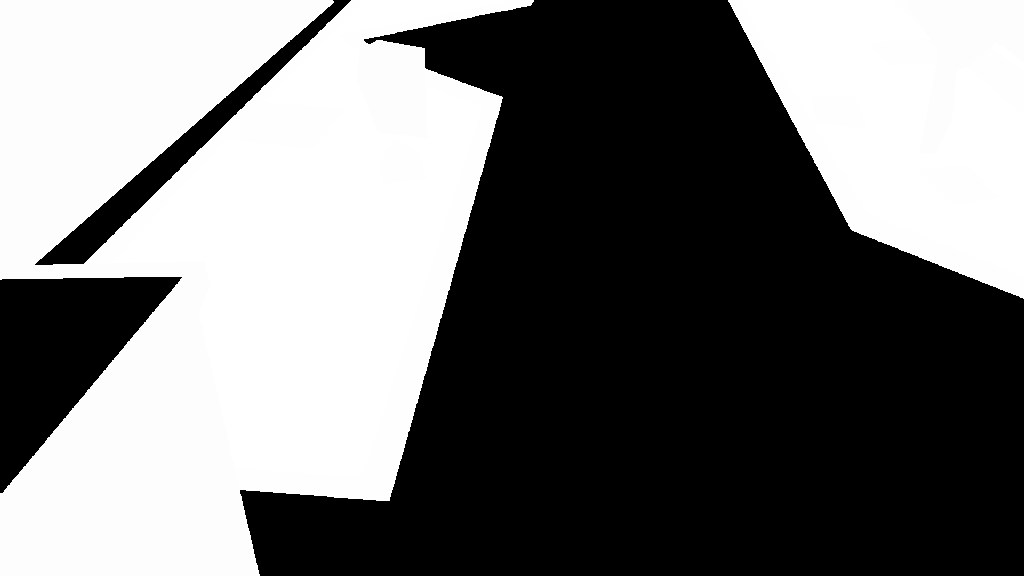}
        \caption*{}
      \end{subfigure}
      \hfill
      \begin{subfigure}[b]{0.20\linewidth}
        \centering
        \includegraphics[width=\linewidth]{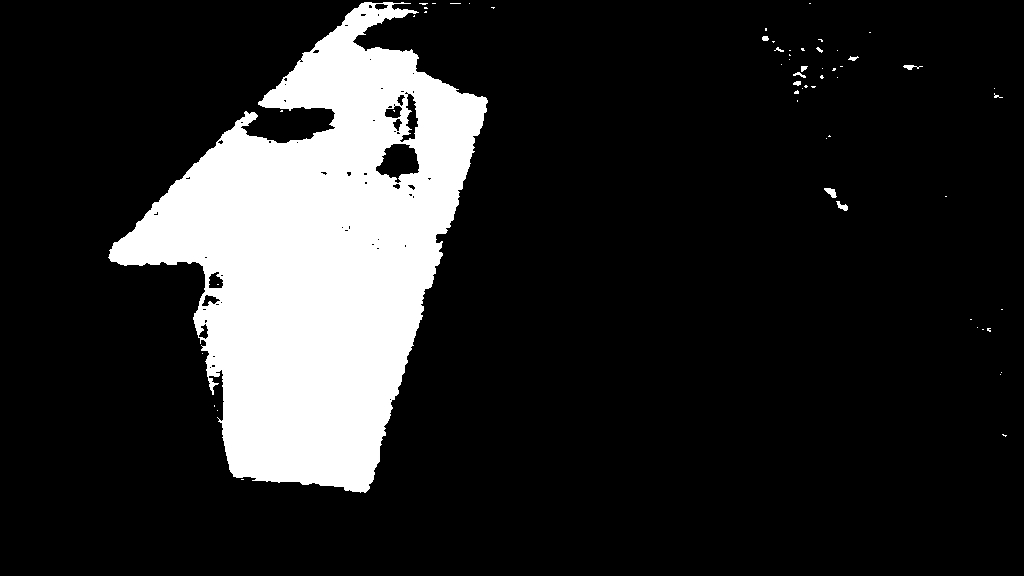}
        \caption*{}
      \end{subfigure}
      \hfill
      \begin{subfigure}[b]{0.20\linewidth}
        \centering
        \includegraphics[width=\linewidth]{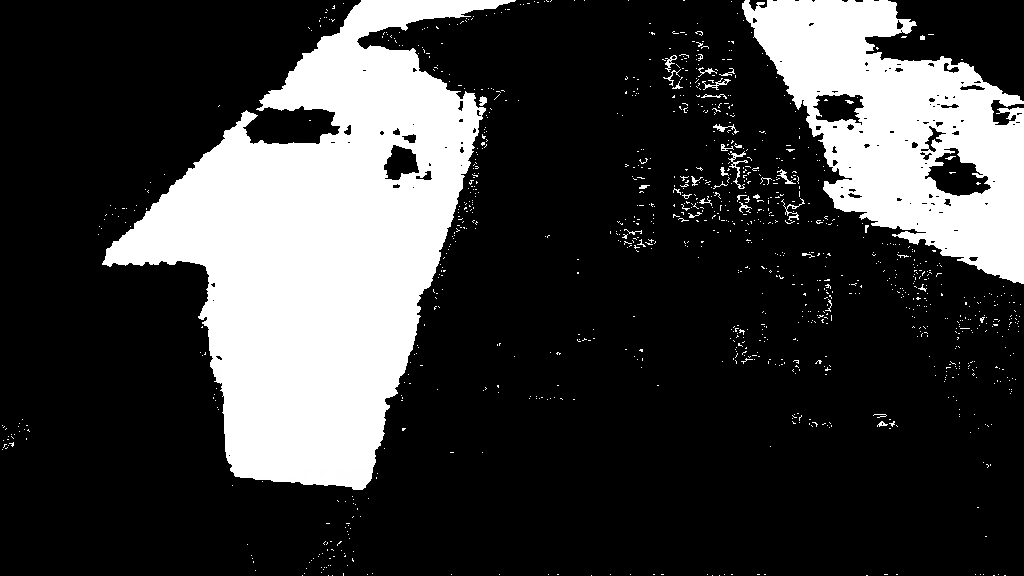}
        \caption*{}
      \end{subfigure} 

    \vspace{-4mm}
      \begin{subfigure}[b]{0.20\linewidth}
        \centering
        \includegraphics[width=\linewidth]{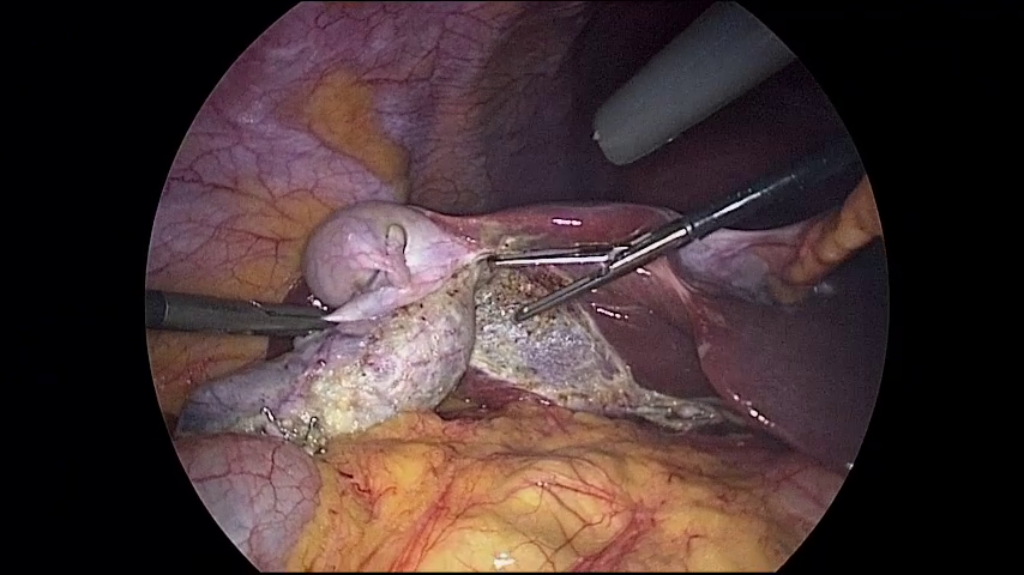}
        \caption*{}
      \end{subfigure}
      \hfill
      \begin{subfigure}[b]{0.20\linewidth}
        \centering
        \includegraphics[width=\linewidth]{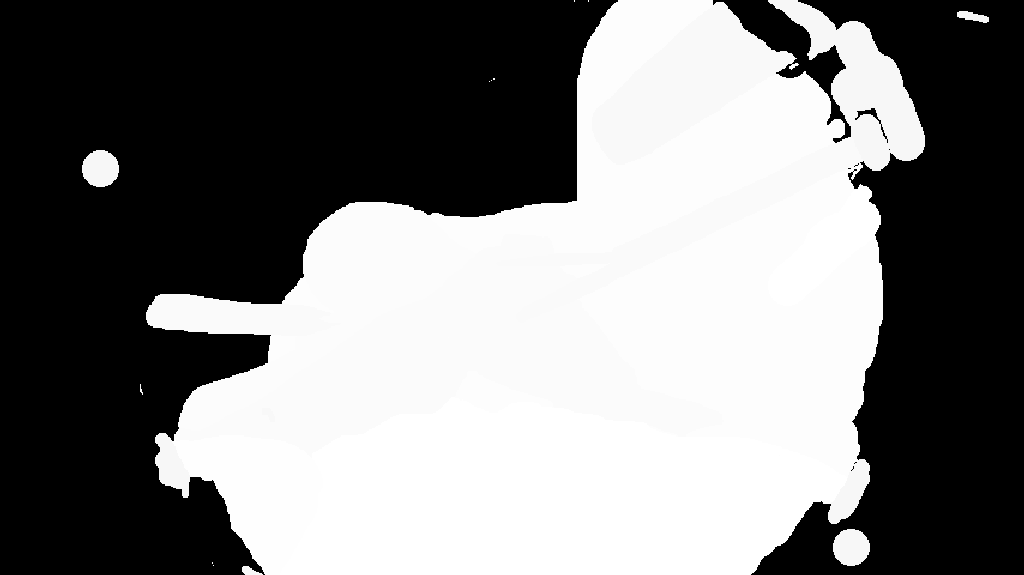}
        \caption*{}
      \end{subfigure}
      \hfill
      \begin{subfigure}[b]{0.20\linewidth}
        \centering
        \includegraphics[width=\linewidth]{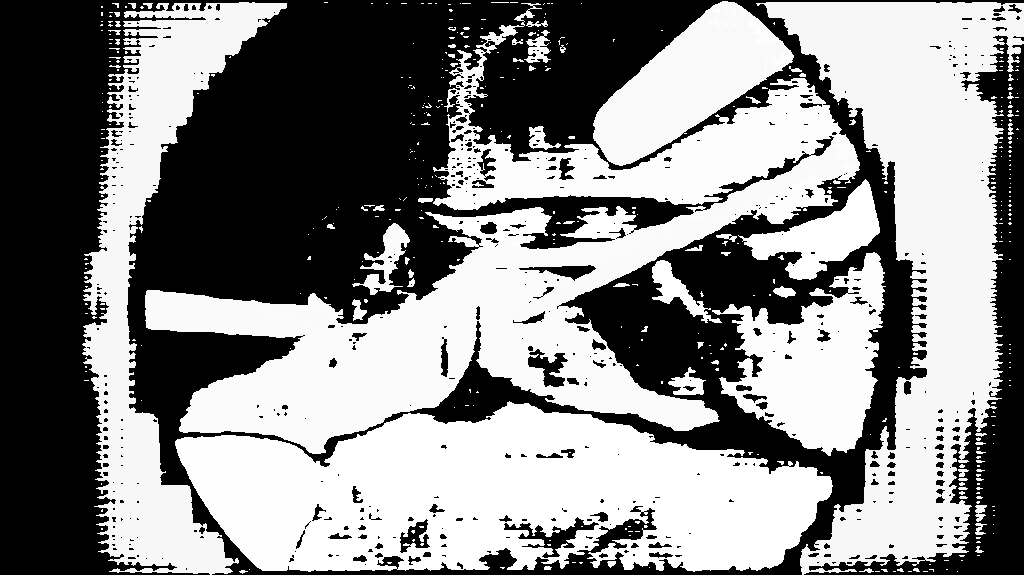}
        \caption*{}
      \end{subfigure}
      \hfill
      \begin{subfigure}[b]{0.20\linewidth}
        \centering
        \includegraphics[width=\linewidth]{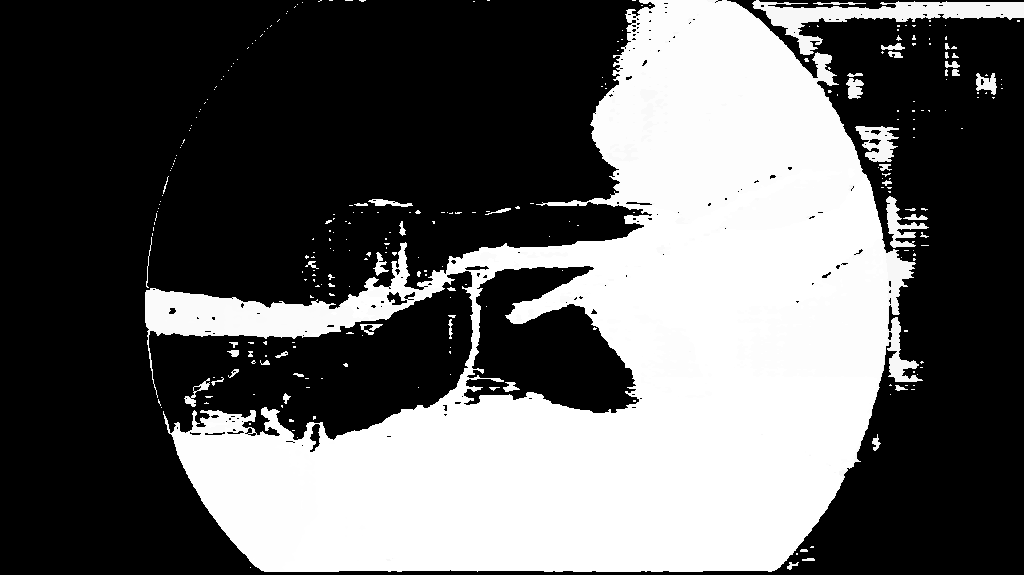}
        \caption*{}
      \end{subfigure} 
      
    \end{minipage}
  }
    \caption{
    \textbf{Comparison of segmentation results across Multiclass datasets}: QTT-SEG demonstrates efficient domain generalization under 60 seconds of tuning. All labels are white and the background is black. Order top to bottom: US, human\_parsing, golf, terrain, cholec.
    }
    \label{fig:multi-predictions_grid}
\end{figure}

\begin{figure}[ht]
\setlength{\belowcaptionskip}{-5pt}
  \centering
  \resizebox{0.8\linewidth}{!}{ %
    \begin{minipage}{\textwidth} %

    \begin{subfigure}[b]{0.20\linewidth}
      \centering
      \makebox[0pt]{\textbf{Input}}
    \end{subfigure}
    \hfill
    \begin{subfigure}[b]{0.20\linewidth}
      \centering
      \makebox[0pt]{\textbf{Ground Truth}}
    \end{subfigure}
    \hfill
    \begin{subfigure}[b]{0.20\linewidth}
      \centering
      \makebox[0pt]{\textbf{SAM-Zero-Shot}}
    \end{subfigure}
    \hfill
    \begin{subfigure}[b]{0.20\linewidth}
      \centering
      \makebox[0pt]{\textbf{QTT-SEG}}
    \end{subfigure}
    
    \vspace{1mm}
      \centering
      \begin{subfigure}[b]{0.20\linewidth}
        \centering
        \includegraphics[width=\linewidth]{images/preds/polyp/image_2.png}
        \caption*{}
      \end{subfigure}
      \hfill
      \begin{subfigure}[b]{0.20\linewidth}
        \centering
        \includegraphics[width=\linewidth]{images/preds/polyp/gt_mask_2.png}
        \caption*{}
      \end{subfigure}
      \hfill
      \begin{subfigure}[b]{0.20\linewidth}
        \centering
        \includegraphics[width=\linewidth]{images/preds/polyp/prd_mask_2_zero.png}
        \caption*{}
      \end{subfigure}
      \hfill
      \begin{subfigure}[b]{0.20\linewidth}
        \centering
        \includegraphics[width=\linewidth]{images/preds/polyp/prd_mask_2.png}
        \caption*{}
      \end{subfigure}

      \vspace{-4mm}

      \begin{subfigure}[b]{0.20\linewidth}
        \centering
        \includegraphics[width=\linewidth]{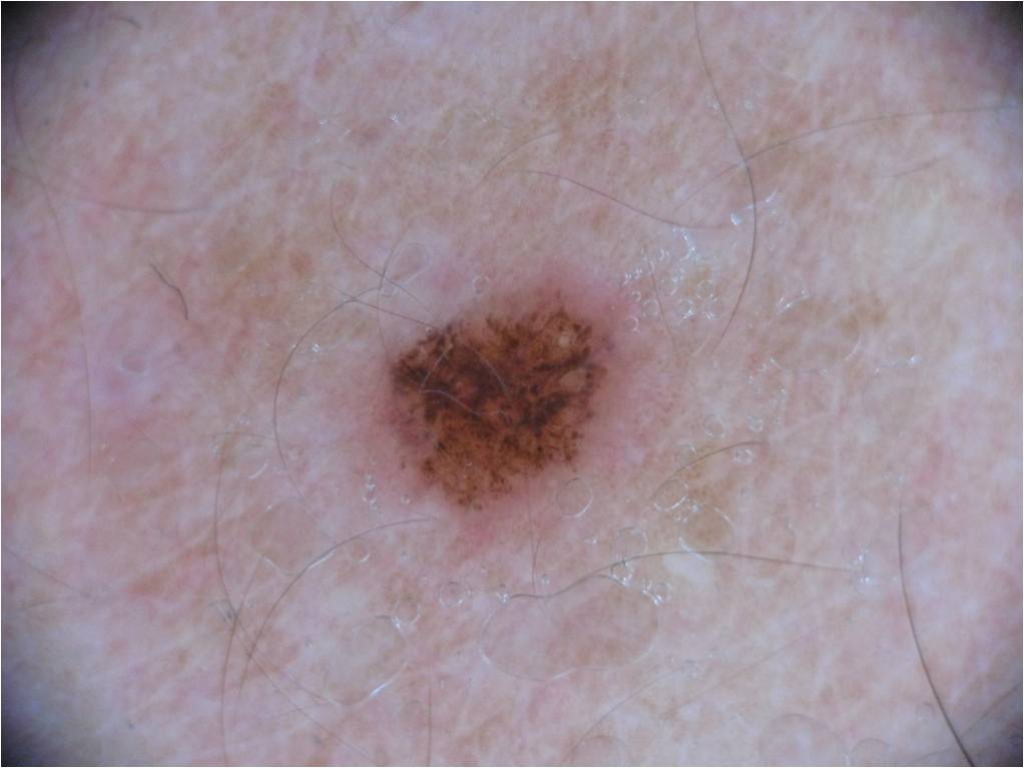}
        \caption*{}
      \end{subfigure}
      \hfill
      \begin{subfigure}[b]{0.20\linewidth}
        \centering
        \includegraphics[width=\linewidth]{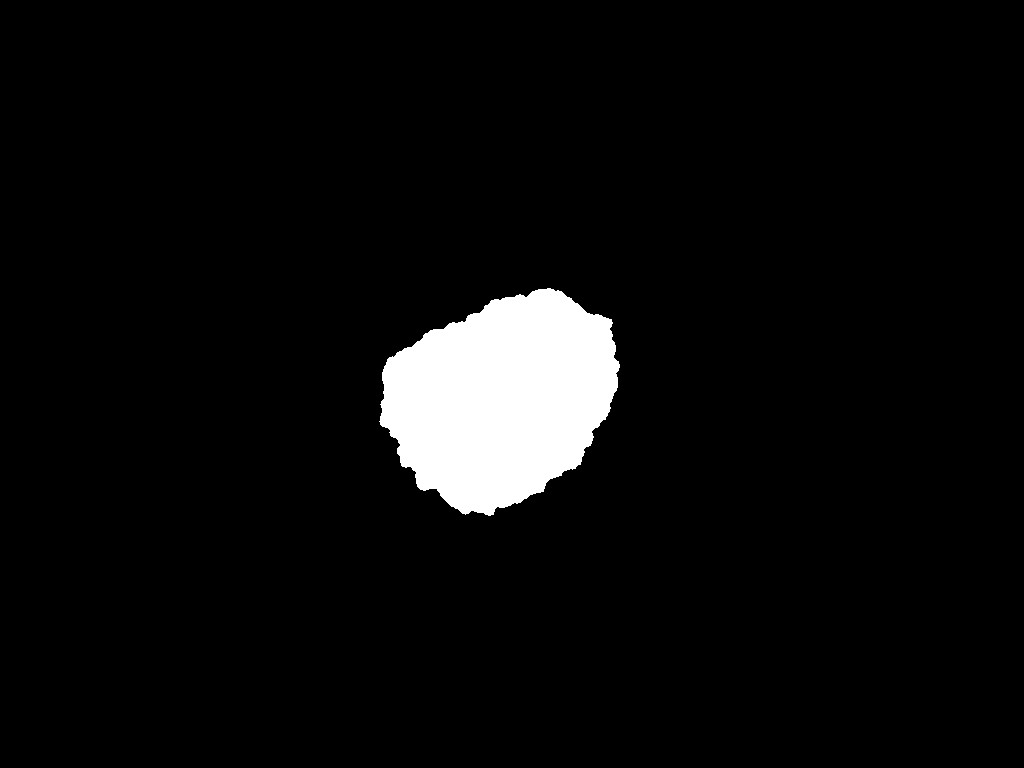}
        \caption*{}
      \end{subfigure}
      \hfill
      \begin{subfigure}[b]{0.20\linewidth}
        \centering
        \includegraphics[width=\linewidth]{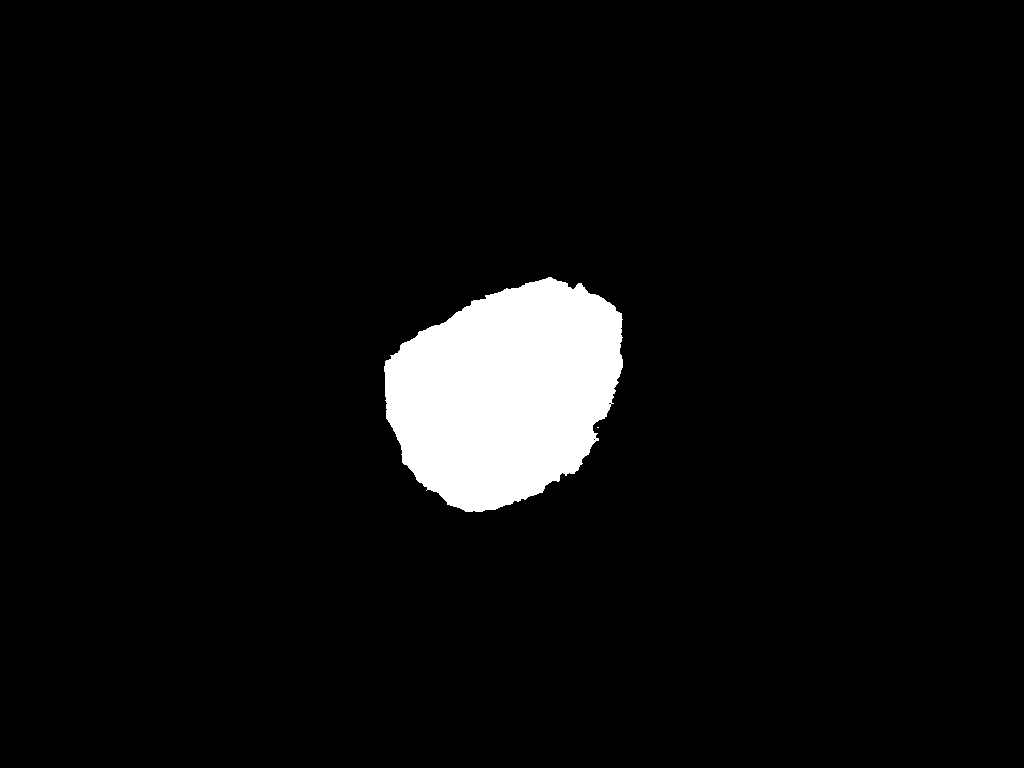}
        \caption*{}
      \end{subfigure}
      \hfill
      \begin{subfigure}[b]{0.20\linewidth}
        \centering
        \includegraphics[width=\linewidth]{images/preds/lesion/prd_mask_1.png}
        \caption*{}
      \end{subfigure}

      \vspace{-4mm}
      
      \begin{subfigure}[b]{0.20\linewidth}
        \centering
        \includegraphics[width=\linewidth]{images/preds/leaf/image_0.png}
        \caption*{}
      \end{subfigure}
      \hfill
      \begin{subfigure}[b]{0.20\linewidth}
        \centering
        \includegraphics[width=\linewidth]{images/preds/leaf/gt_mask_0.png}
        \caption*{}
      \end{subfigure}
      \hfill
      \begin{subfigure}[b]{0.20\linewidth}
        \centering
        \includegraphics[width=\linewidth]{images/preds/leaf/prd_mask_0_zero.png}
        \caption*{}
      \end{subfigure}
      \hfill
      \begin{subfigure}[b]{0.20\linewidth}
        \centering
        \includegraphics[width=\linewidth]{images/preds/leaf/prd_mask_0.png}
        \caption*{}
      \end{subfigure}

    \vspace{-4mm}
    
      \begin{subfigure}[b]{0.20\linewidth}
        \centering
        \includegraphics[width=\linewidth]{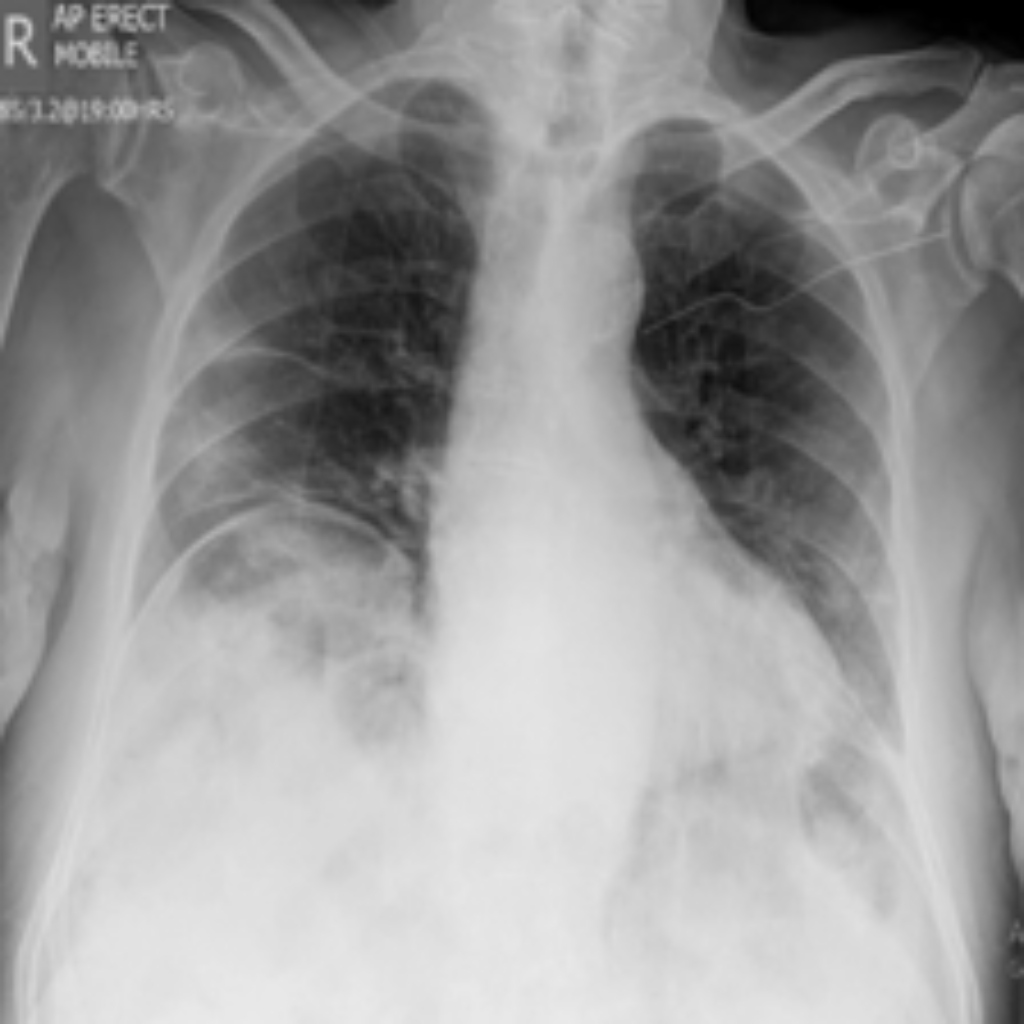}
        \caption*{}
      \end{subfigure}
      \hfill
      \begin{subfigure}[b]{0.20\linewidth}
        \centering
        \includegraphics[width=\linewidth]{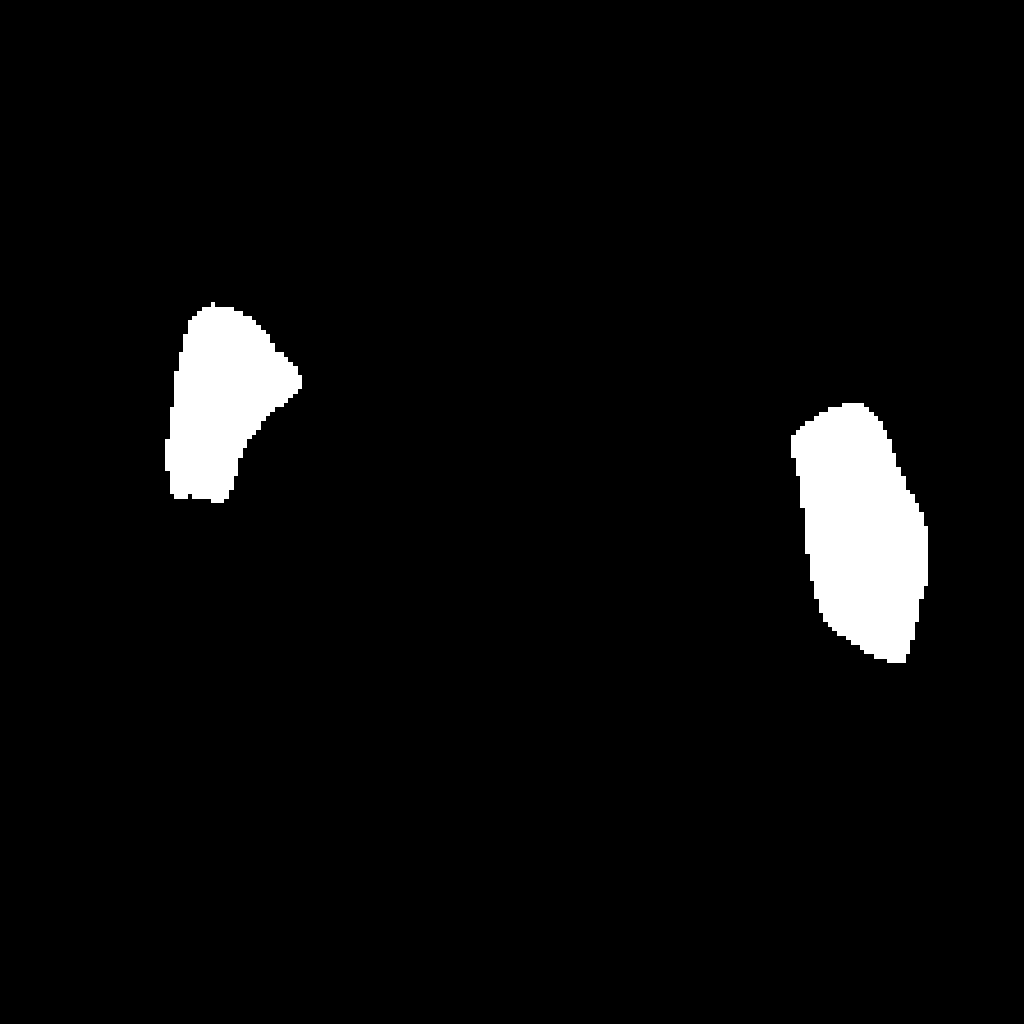}
        \caption*{}
      \end{subfigure}
      \hfill
      \begin{subfigure}[b]{0.20\linewidth}
        \centering
        \includegraphics[width=\linewidth]{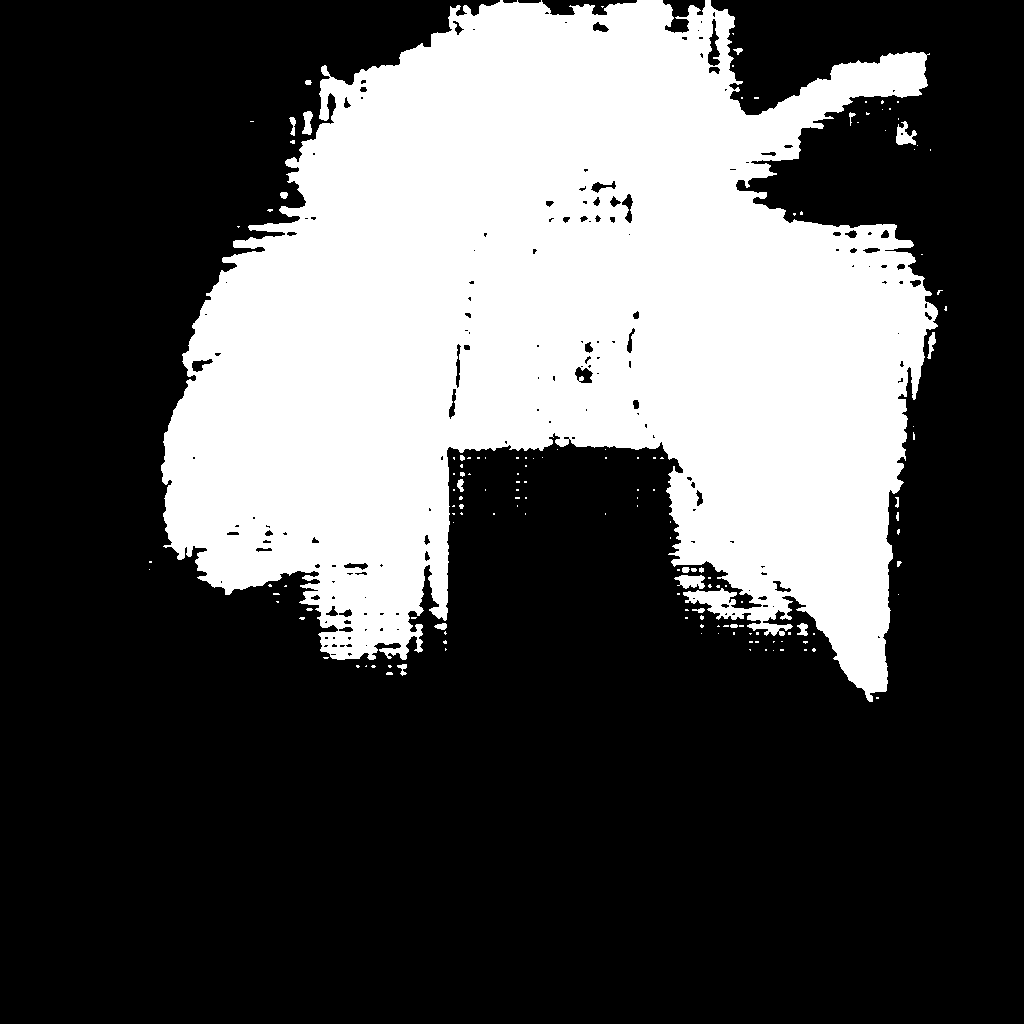}
        \caption*{}
      \end{subfigure}
      \hfill
      \begin{subfigure}[b]{0.20\linewidth}
        \centering
        \includegraphics[width=\linewidth]{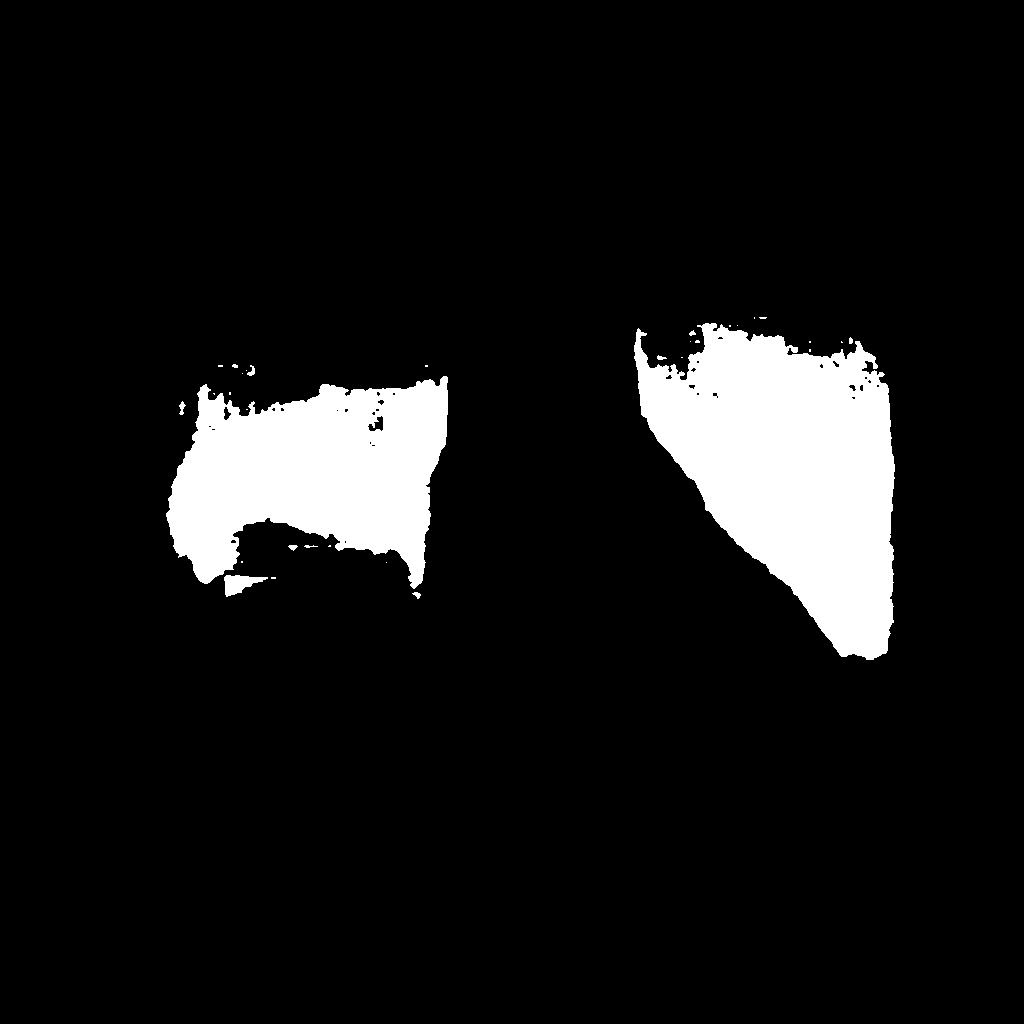}
        \caption*{}
      \end{subfigure}

      \vspace{-4mm}
            \begin{subfigure}[b]{0.20\linewidth}
        \centering
        \includegraphics[width=\linewidth]{images/preds/eyes/image_1.png}
        \caption*{}
      \end{subfigure}
      \hfill
      \begin{subfigure}[b]{0.20\linewidth}
        \centering
        \includegraphics[width=\linewidth]{images/preds/eyes/gt_mask_1.png}
        \caption*{}
      \end{subfigure}
      \hfill
      \begin{subfigure}[b]{0.20\linewidth}
        \centering
        \includegraphics[width=\linewidth]{images/preds/eyes/prd_mask_1_zero.png}
        \caption*{}
      \end{subfigure}
      \hfill
      \begin{subfigure}[b]{0.20\linewidth}
        \centering
        \includegraphics[width=\linewidth]{images/preds/eyes/prd_mask_1.png}
        \caption*{}
      \end{subfigure}

      \vspace{-4mm}
            \begin{subfigure}[b]{0.20\linewidth}
        \centering
        \includegraphics[width=\linewidth]{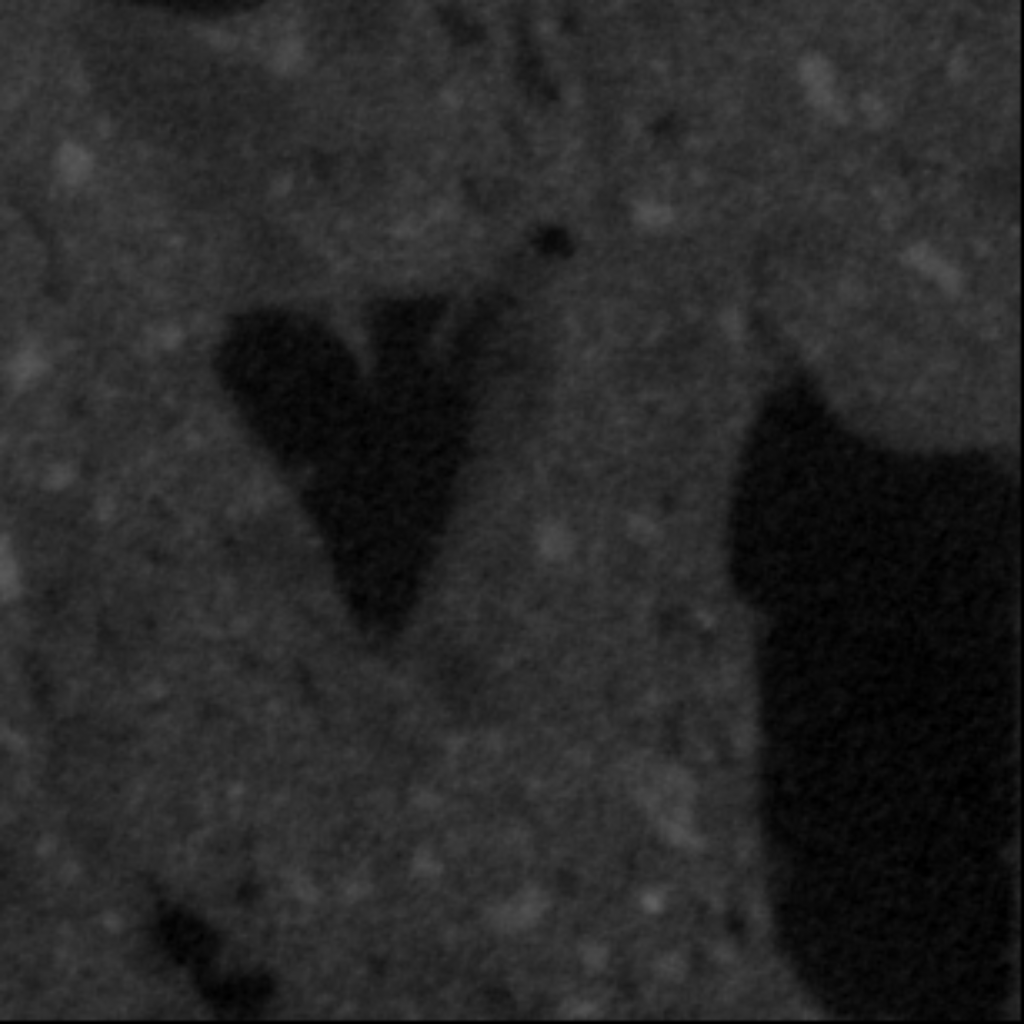}
        \caption*{}
      \end{subfigure}
      \hfill
      \begin{subfigure}[b]{0.20\linewidth}
        \centering
        \includegraphics[width=\linewidth]{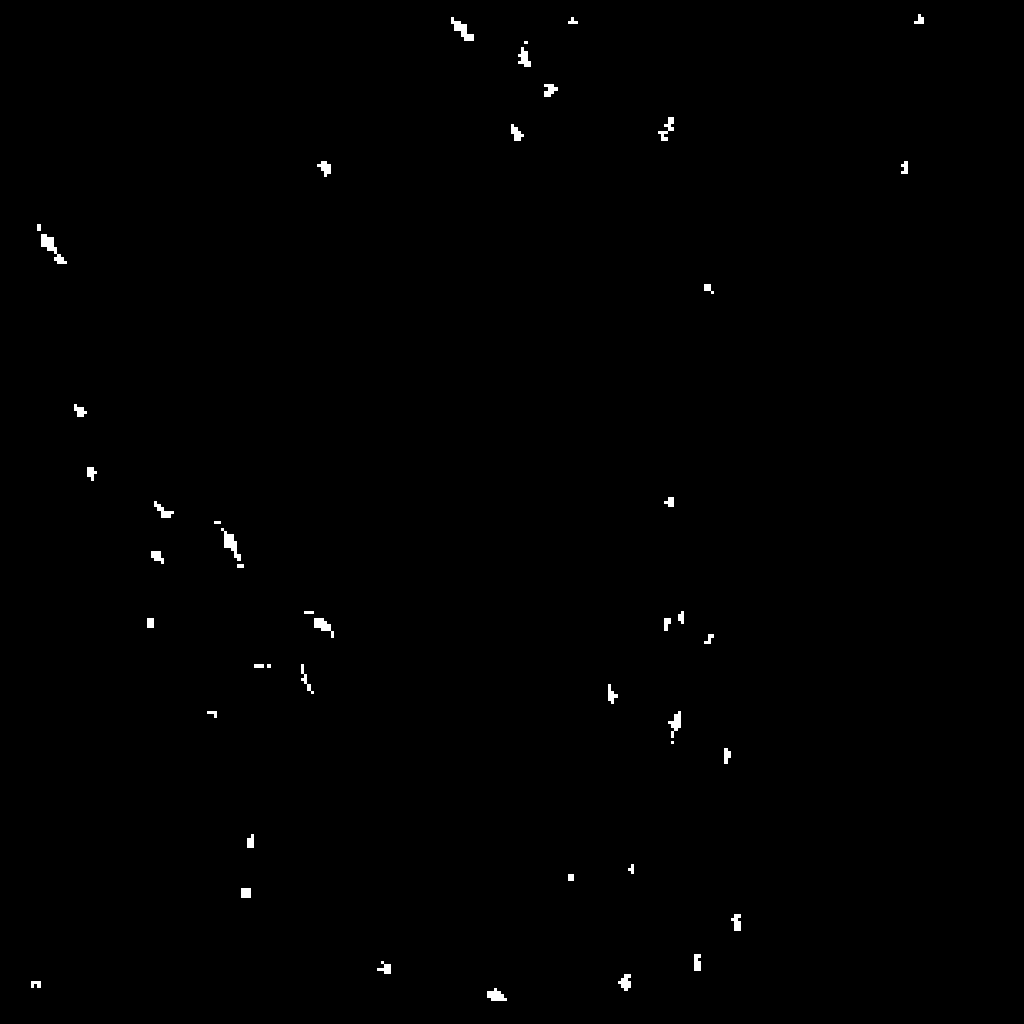}
        \caption*{}
      \end{subfigure}
      \hfill
      \begin{subfigure}[b]{0.20\linewidth}
        \centering
        \includegraphics[width=\linewidth]{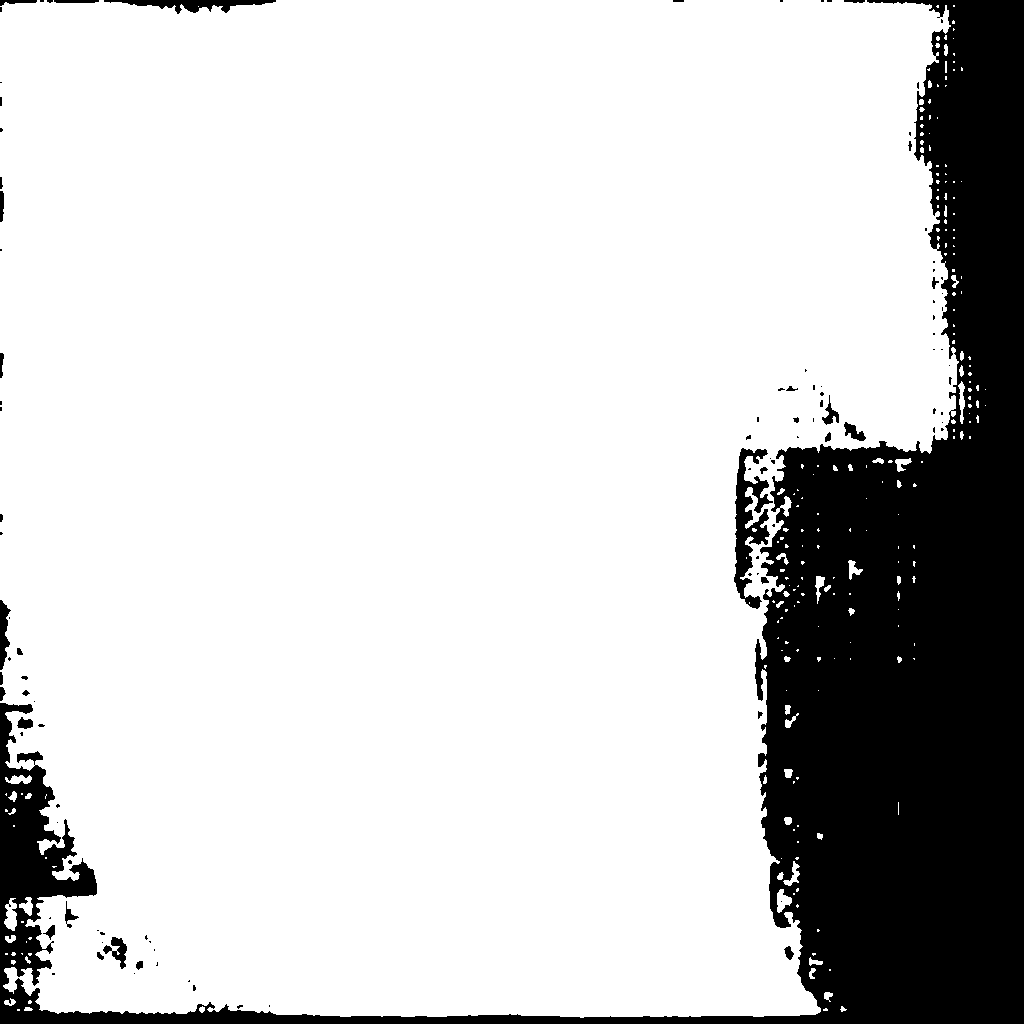}
        \caption*{}
      \end{subfigure}
      \hfill
      \begin{subfigure}[b]{0.20\linewidth}
        \centering
        \includegraphics[width=\linewidth]{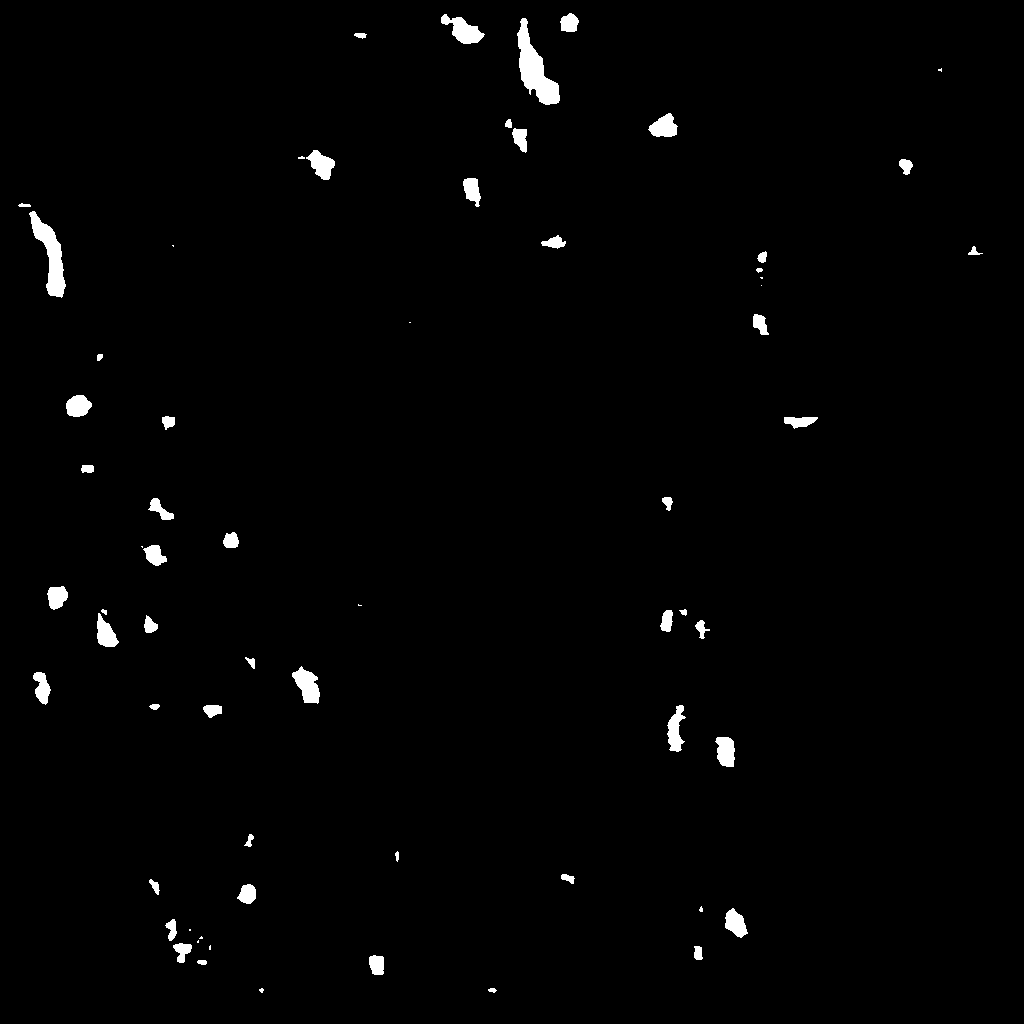}
        \caption*{}
      \end{subfigure}

      \vspace{-4mm}
            \begin{subfigure}[b]{0.20\linewidth}
        \centering
        \includegraphics[width=\linewidth]{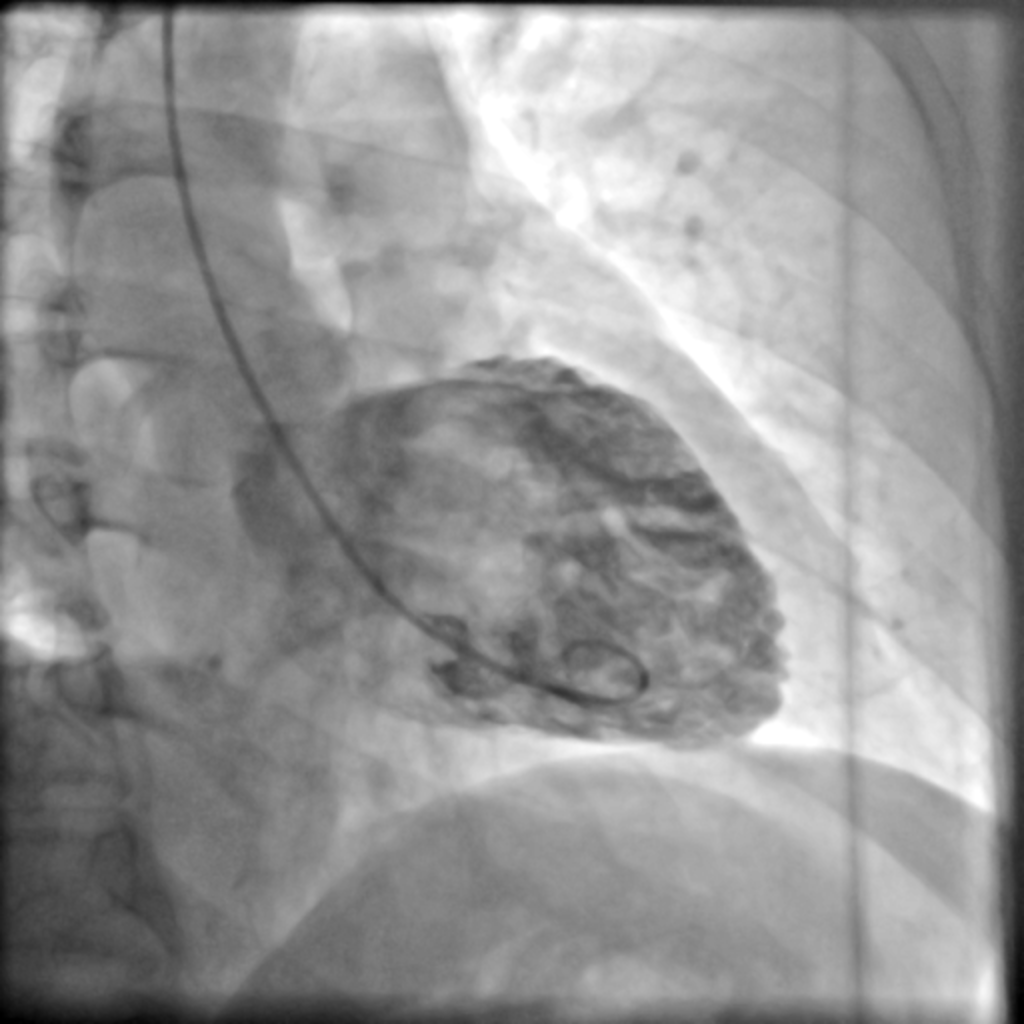}
        \caption*{}
      \end{subfigure}
      \hfill
      \begin{subfigure}[b]{0.20\linewidth}
        \centering
        \includegraphics[width=\linewidth]{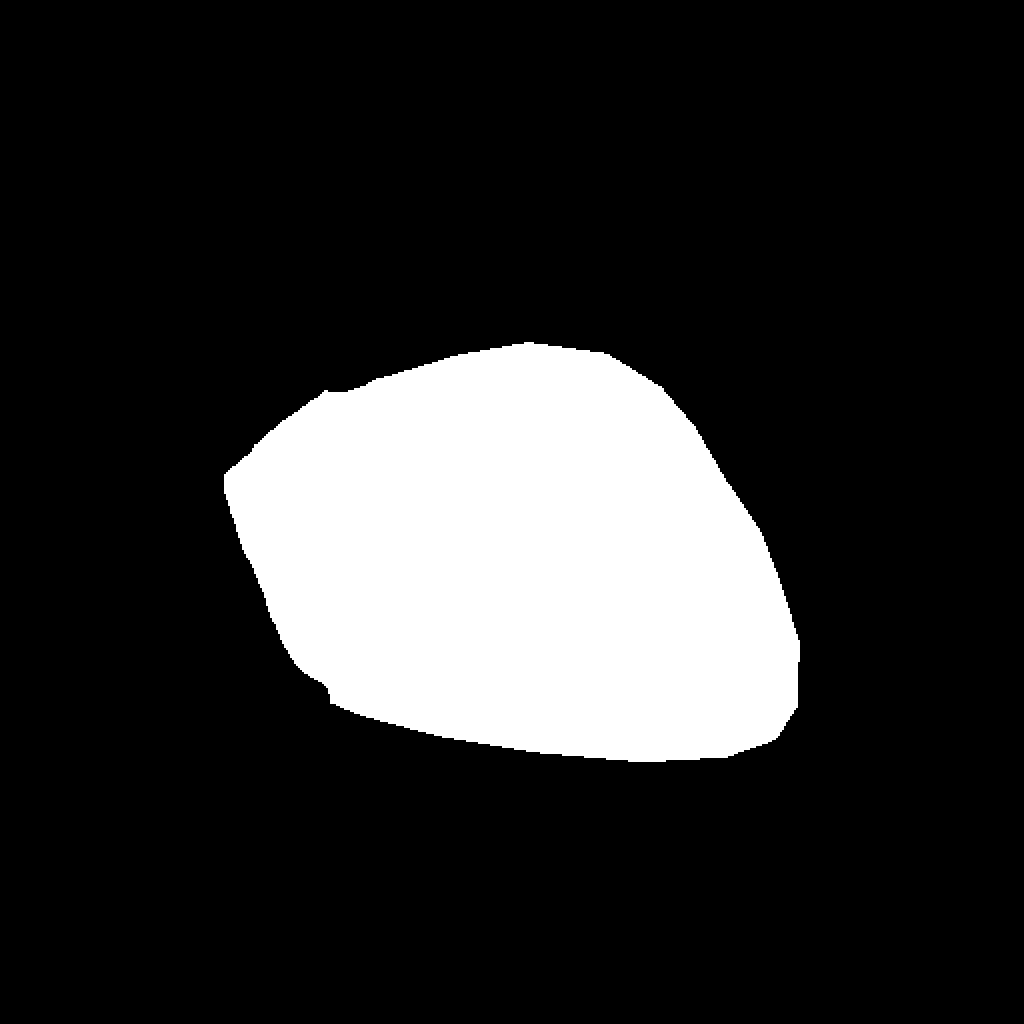}
        \caption*{}
      \end{subfigure}
      \hfill
      \begin{subfigure}[b]{0.20\linewidth}
        \centering
        \includegraphics[width=\linewidth]{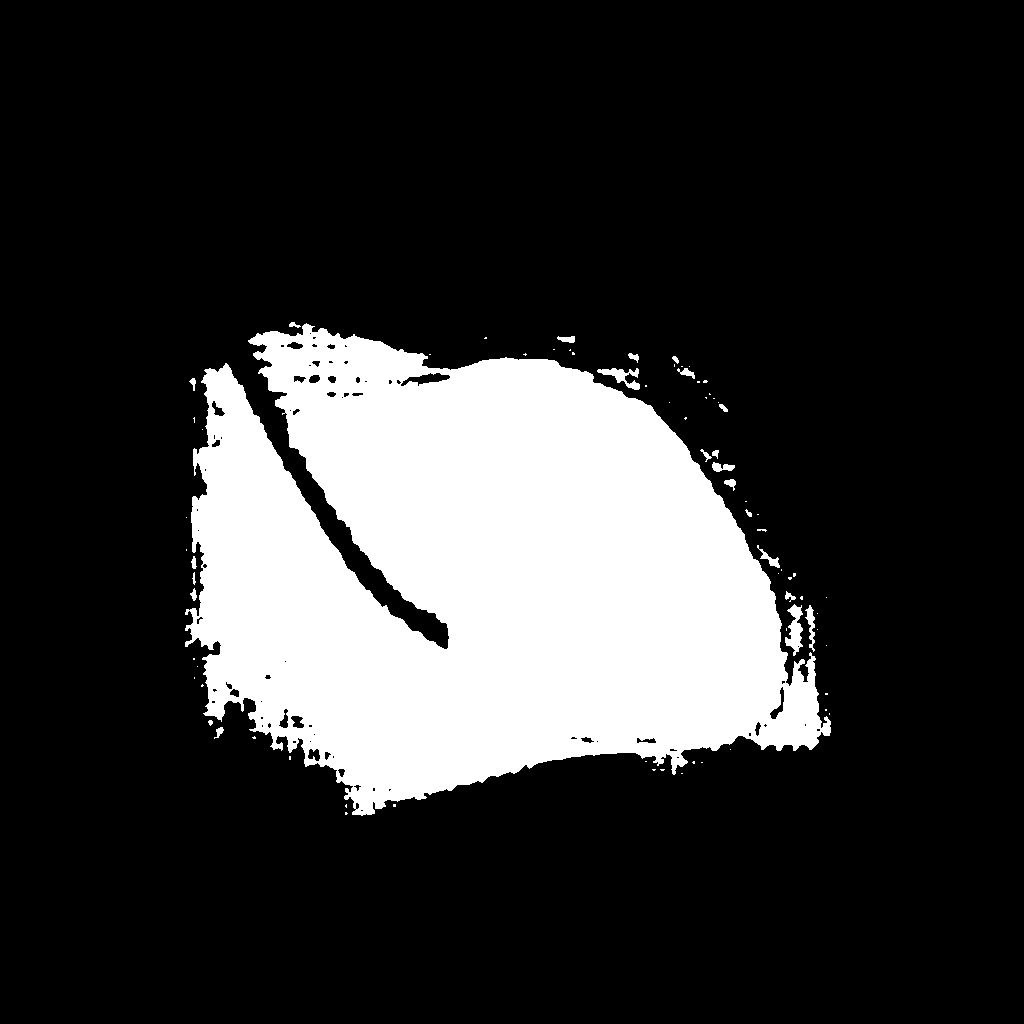}
        \caption*{}
      \end{subfigure}
      \hfill
      \begin{subfigure}[b]{0.20\linewidth}
        \centering
        \includegraphics[width=\linewidth]{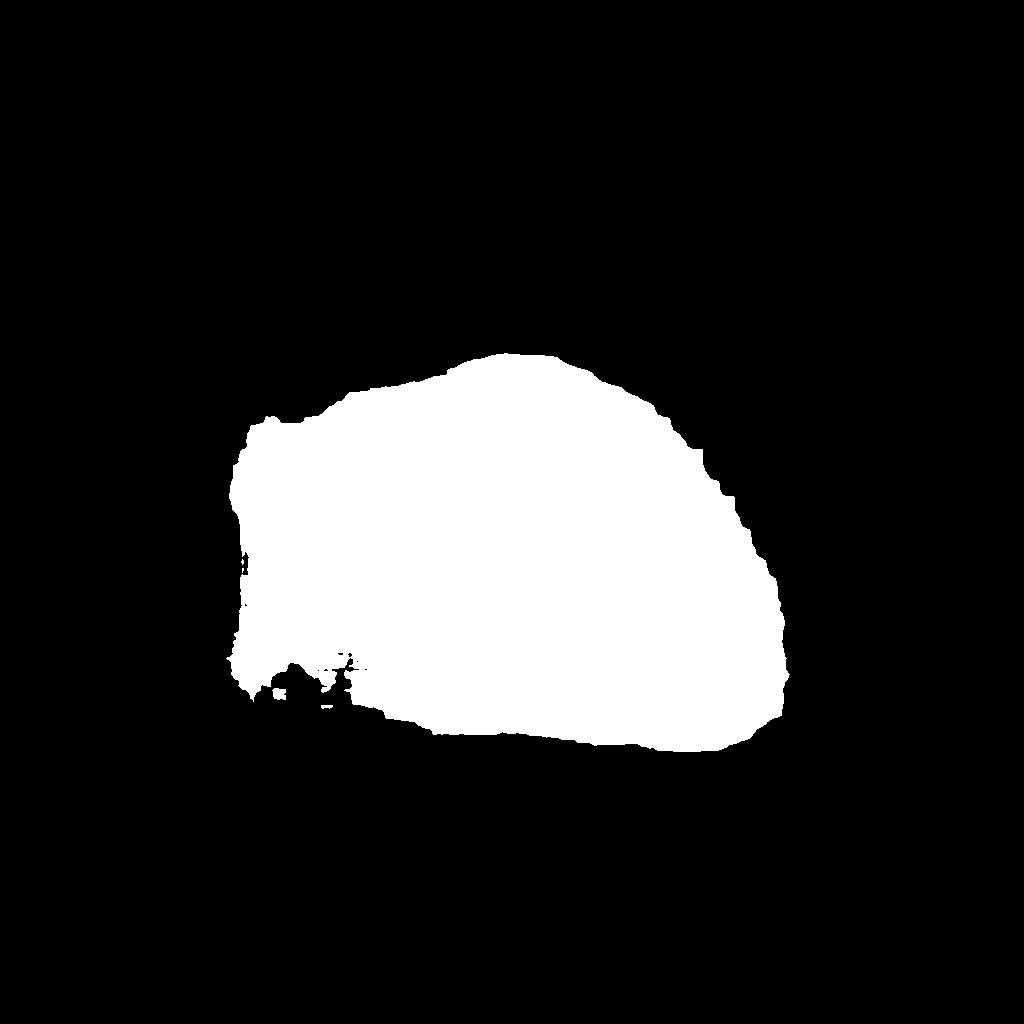}
        \caption*{}
      \end{subfigure}
      
      \vspace{-4mm}
      \begin{subfigure}[b]{0.20\linewidth}
        \centering
        \includegraphics[width=\linewidth]{images/preds/chest/image_1.png}
        \caption*{}
      \end{subfigure}
      \hfill
      \begin{subfigure}[b]{0.20\linewidth}
        \centering
        \includegraphics[width=\linewidth]{images/preds/chest/gt_mask_1.png}
        \caption*{}
      \end{subfigure}
      \hfill
      \begin{subfigure}[b]{0.20\linewidth}
        \centering
        \includegraphics[width=\linewidth]{images/preds/chest/prd_mask_1_zero.png}
        \caption*{}
      \end{subfigure}
      \hfill
      \begin{subfigure}[b]{0.20\linewidth}
        \centering
        \includegraphics[width=\linewidth]{images/preds/chest/prd_mask_1.png}
        \caption*{}
      \end{subfigure}
    \end{minipage}
  }
    \caption{
    \textbf{Comparison of segmentation results across binary datasets}: QTT-SEG demonstrates efficient domain generalization under 60 seconds of tuning. All labels are white and the background is black. Order from top to bottom: polyp, lesion, leaf, covid, eyes, fiber, cardiac, chest.
    }
    \label{fig:binary-predictions_grid}
\end{figure}

\end{document}